\documentclass{article} % For LaTeX2e
\usepackage[final]{colm2025_conference}

\usepackage{microtype}
\usepackage{hyperref}
\usepackage{url}
\usepackage{booktabs}

\usepackage{lineno}

\definecolor{darkblue}{rgb}{0, 0, 0.5}
\hypersetup{colorlinks=true, citecolor=darkblue, linkcolor=darkblue, urlcolor=darkblue}

\usepackage[utf8]{inputenc} % allow utf-8 input
\usepackage[T1]{fontenc}    % use 8-bit T1 fonts
\usepackage{amsfonts}       % blackboard math symbols
\usepackage{nicefrac}       % compact symbols for 1/2, etc.

\usepackage{amsmath}
\usepackage{amsthm}
\usepackage{amssymb}
\usepackage[capitalise]{cleveref}
\usepackage{colortbl}
\usepackage{multirow}
\usepackage{siunitx}
\usepackage{algorithm}
\usepackage{algpseudocode}
\usepackage[title]{appendix}
\usepackage{wrapfig}
\usepackage{enumerate}
\usepackage{etoolbox}
\AtBeginEnvironment{tcolorbox}{\footnotesize}

\usepackage{thmtools}
\usepackage{thm-restate}

\DeclareMathOperator*{\argmax}{argmax}
\DeclareMathOperator*{\argmin}{argmin}

\usepackage[bottom]{footmisc}

\hypersetup{
  colorlinks,
  linkcolor={blue!50!black},
  citecolor={blue!50!black},
  urlcolor={blue!50!black}
}

\usepackage{algpseudocode}

\usepackage{amsmath,amsfonts,bm}

\def\1{\bm{1}}

\def\rvx{{\mathbf{x}}}
\def\rvy{{\mathbf{y}}}

\def\vx{{\bm{x}}}
\def\vy{{\bm{y}}}

\DeclareMathAlphabet{\mathsfit}{\encodingdefault}{\sfdefault}{m}{sl}
\SetMathAlphabet{\mathsfit}{bold}{\encodingdefault}{\sfdefault}{bx}{n}

\newcommand{\R}{\mathbb{R}}
\newcommand{\E}{\mathop{\mathbb{E}}}
\newcommand{\D}{\mathcal{D}}

\newcommand{\abs}[1]{\vert #1 \vert}

\newcommand{\piref}{\pi_\text{ref}}
\newcommand{\score}{\mathrm{score}}

\newcommand{\KL}{D_{\mathrm{KL}}}
\newcommand{\softmax}{\mathrm{softmax}}

\theoremstyle{definition}

\usepackage[most]{tcolorbox}

\title{A Critical Look At \\ Tokenwise Reward-Guided Text Generation}

\author{%
  Ahmad Rashid \\
  University of Waterloo \\ Vector Institute
  \And
  Ruotian Wu \\
  University of Waterloo \\
  Vector Institute \\
  \And
  \And
  \AND
  Julia Grosse \\
  University of T\"{u}bingen \\
  T\"{u}bingen AI Center \\
  \And
  Agustinus Kristiadi\thanks{Equal Supervision} \\
  Western University \\
  Vector Institute \\
  \And
  Pascal Poupart$^{*}$ \\
  University of Waterloo \\ Vector Institute \\
}

\begin{document}

\ifcolmsubmission
\linenumbers
\fi

\maketitle

\begin{abstract}
  Large language models (LLMs) can be improved by aligning with human preferences through fine-tuning---the so-called reinforcement learning from human feedback (RLHF).
  However, the cost of fine-tuning an LLM is prohibitive for many users.
  Due to their ability to bypass LLM fine-tuning, prediction-time tokenwise reward-guided text generation (RGTG) methods have recently been proposed.
  They use a reward model trained on full sequences to score partial sequences during decoding in a bid to steer the generation towards sequences with high rewards.
  However, these methods have so far been only heuristically motivated and poorly analyzed.
  In this work, we show that reward models trained on full sequences are not compatible with scoring partial sequences.
  To alleviate this, we propose to train a Bradley-Terry reward model on partial sequences explicitly, and autoregressively sample from the implied tokenwise policy during decoding.
  We study the properties of this reward model and the resulting policy:
we show that this policy is proportional to the ratio of two distinct RLHF policies.
  Our simple approach outperforms previous RGTG methods and performs similarly to strong offline baselines without large-scale LLM fine-tuning. Code for our work is available at https://github.com/ahmadrash/PARGS
\end{abstract}

\section{Introduction}
\label{sec:intro}
%!TEX root=iclr2025_conference.tex

Large language models (LLMs) provide a modern foundation for most, if not all, text generation tasks \citep{radford2019gpt2,brown2020gpt3,touvron2023llama,touvron2023llama2}.
In practice, significant improvements in the quality of text generation are achieved by aligning LLMs to human preferences \citep{stiennon2020rlhf,ouyang2022training}.
This is typically performed via reinforcement learning from human feedback (RLHF), which involves two steps: i) learning a reward model from preference data and ii) fine-tuning an LLM to maximize expected rewards by reinforcement learning \citep{ziegler2019rlhf}.
Usually, this is done via a reinforcement learning algorithm such as proximal policy optimization \citep[PPO,][]{schulman2017proximal}.
Nevertheless, recently, \citet{rafailov2023direct} showed that the reward modeling step (i) can be bypassed by directly fine-tuning an LLM with preference data, resulting in a method called direct preference optimization (DPO).
Although this simplifies RLHF, the fine-tuning step (ii) remains prohibitively costly for most users, since it requires high-performance computational resources with large GPUs.

In order to alleviate the computational issue above, \citet{khanov2023alignment,deng2023reward} explored tokenwise reward-guided text generation (RGTG) techniques that avoid any fine-tuning of the LLM.
More precisely, the LLM remains frozen (i.e., not fine-tuned), and the reward model is used at the decoding time to adjust the softmax scores of the tokens directly.
Unlike DPO, this line of work retains the reward modeling step, but training reward models is typically a much cheaper endeavor than fine-tuning text-generation LLMs since smaller models can be utilized for reward modeling.
Furthermore, reward models are modular: they can easily be composed and reused without any cost to guide text generation in conjunction with any base LLM.
In contrast, RLHF via DPO requires fine-tuning of every LLM that we wish to improve based on human preference data.

While RGTG is an interesting alternative to the standard offline RLHF, it is often based on heuristics and still poorly analyzed.
For instance, ARGS \citep{khanov2023alignment} proposed to simply use a reward model trained on full sequences to score each partial sequence during autoregressive decoding.
Meanwhile, \citet{deng2023reward} used a custom tokenwise loss to distill a reward model trained on full sequences. Thus, it is unclear whether these approaches can give rise to a sound tokenwise text generation policy. Controlled decoding \citep[CD;][]{mudgalcontrolled}, on the other hand, uses rollouts from the base model along with a reward model trained on full sequences to distill the partial reward.

In this work, we analyze this common RGTG approach.
First, we show that the usage of full-sequence reward models to score partial sequences in a tokenwise policy is pathological.
To alleviate this, we propose to explicitly train a Bradley-Terry (BT) reward model on partial sequences.
We prove that this text generation policy is a ratio of two different RLHF policies trained on sequences of different lengths.
Ideally, the policy would be derived from a single RLHF policy, but as we shall also show in \cref{sec:method}, the exact computation of such a policy is intractable.
By deriving the policy from a ratio of distinct RLHF policies, we obtain a tractable sampling procedure.
This is akin to the argument of \citet{zhao2024probabilistic} where they use a similar ratio to derive a sequential Monte Carlo method.

We empirically validate our analysis on three different text generation datasets on two recent LLMs.
Evidence shows that our RGTG approach achieves better performance compared to ARGS and CD, matching the performance of the more expensive, offline PPO and DPO baselines.
In summary:
\begin{enumerate}[(i)]
  \vspace{-0.5em}
  \setlength\itemsep{0em}
  \item We analyze the recent practice of using full-sequence reward models for guiding the LLM decoding process. In particular, we show a deficiency in this approach.
  \item We thus propose to explicitly train a BT reward model on partial sequences and sample from the induced per-token policy induced by it during the decoding time.
  \item We show that this reward model induces a ratio of two distinct RLHF policies over sequences with different lengths. This is a trade-off that one must make to make tokenwise RGTG free of the aforementioned deficiency and yet still tractable.
  \item Extensive experiments with multiple LLMs on text generation validate our insights.
\end{enumerate}

\section{Preliminaries}
\label{sec:prelim}
%!TEX root=iclr2025_conference.tex

% We will use the following notation throughout the paper.
We denote a prompt by $\rvx$ and its response by $\rvy$ where the bolded letters indicate sequences of tokens.
The $i$-th token in $\rvx$ is denoted by $x^i$, while the partial sequence starting at token $i$ and ending at token $j$ is denoted by $\rvx^{i:j}$.  The length of a sequence $\rvx$ is denoted by $\abs{\rvx}$.
The same notation applies to \(\rvy\).

\subsection{Reinforcement Learning from Human Feedback}

LLMs generally consist of probabilistic models that can generate a response $\rvy$ given a prompt $\rvx$.
More specifically, the generation of $\rvy$ is done token-by-token by sampling the next token from a conditional distribution $\pi(y^{i} | \rvx,\rvy^{1:i-1})$.

Given a preference dataset \(\D = \{ (\rvx_k, \rvy_{wk}, \rvy_{lk}) \}_{k=1}^{K}\) containing \(K\) triples of token sequences \((\rvx, \rvy_w, \rvy_l)\), \citet{ziegler2019rlhf} and \citet{ouyang2022training} proposed a technique based on reinforcement learning (RL) to align an LLM with the preference dataset.
They train a parametric reward model $r_{\phi}(\rvy|\rvx)$ that assigns a higher score to the ``winning'' (i.e., preferred) utterance $\rvy_w$ and a lower score to the ``losing'' utterance $\rvy_l$.
This is done via the BT model \citep{bradleyterry1952paired} which minimizes the loss:
\begin{equation} \label{eq:bradley-terry}
  \mathcal{L}_R = - \E_{\rvx, \rvy_w, \rvy_l \sim \D} \log \sigma ( r_{\phi} (\rvy_w|\rvx) - r_{\phi} (\rvy_l|\rvx)) ,
\end{equation}
where \(\sigma\) is the logistic function.
Note that \(r_\phi\) is trained to score entire utterances \(\rvy\).
Once $r_\phi$ is trained, it can be used to infer the probability of generating sequence \(\rvy\) in response to \(\rvx\), i.e., \(P_\phi(\rvy|\rvx)=\nicefrac{\exp(r_\phi(\rvy|\rvx))}{\sum_{\rvy'} \exp(r_\phi(\rvy'|\rvx)}\).
Given a reference LLM, we denote by $\piref(\rvy|\rvx)$ the conditional probability that it will generate response $\rvy$ to prompt $\rvx$ (also referred to as policy).
We refer to the LLM and its policy interchangeably.
One can then copy the LLM and fine-tune it to maximize
\begin{equation} \label{eq:rlhf}
  \max_\theta \E_{\substack{\rvx \sim \D, \\ \rvy \sim \pi_\theta(\rvy | \rvx)}}[r_\phi(\rvy \vert \rvx)] - {\frac{1}{\beta}} \KL[\pi_\theta(\rvy | \rvx) \,\Vert\, \piref(\rvy | \rvx)] ,
\end{equation}
where the KL term forms a regularizer that ensures that the fine-tuned model will not differ too much from the reference model.
The optimization problem above can be solved by many RL techniques, including the popular PPO algorithm \citep{schulman2017proximal}.
This RL optimization is quite costly in practice due to the size of the LLM.

% \subsection{Direct Preference Optimization}

The optimization \eqref{eq:rlhf} has a closed form solution of the form \citep{peters2007reinforcement}
\begin{equation} \label{eq:policy}
  \pi_\theta(\rvy | \rvx) = \frac{1}{Z(\rvx)} \piref(\rvy \vert \rvx) \exp(\beta r_\phi(\rvy \vert \rvx))
\end{equation}
where $Z(\rvx) = \sum_\rvy \piref(\rvy | \rvx) \exp(\beta r_\phi(\rvy \vert \rvx))$ is the intractable partition function.
Notice that we can reorganize \eqref{eq:policy} to express the reward function in terms of the policies \(\pi_\theta\) and \(\piref\):
\begin{equation*}
  \label{eq:reward}
  r(\rvy \vert \rvx) = \frac{1}{\beta} \log \frac{\pi_\theta(\rvy | \rvx)}{\piref(\rvy | \rvx)} + \log Z(\rvx) ,
\end{equation*}
which can be used to replace $r_\phi(\rvx|\rvy)$ in \eqref{eq:bradley-terry} to obtain the following optimization problem:
\begin{equation*}
  % \label{eq:dpo}
  \max_\theta \E_{\rvx, \rvy_w, \rvy_l \sim \D} \log \sigma \left( \frac{1}{\beta} \left( \log \frac{\pi_\theta(\rvy_w | \rvx)}{\piref(\rvy_w | \rvx)} - \log \frac{\pi_\theta(\rvy_l| \rvx)}{\piref(\rvy_l| \rvx)}\right) \right) .
\end{equation*}
Maximizing the above objective with respect to $\theta$ directly fine-tunes the LLM without the need to learn a reward model.
Furthermore, this maximization is done by supervised learning, which is generally simpler than RL.
This approach, known as direct preference optimization \citep[DPO,][]{rafailov2023direct}, reduced the cost of RLHF while solving an equivalent optimization problem.
However, note that both PPO and DPO based RLHF are still very costly in practice since they require fine-tuning (a copy of) the target LLM \(\piref\).

\subsection{Reward-Guided Text Generation}
\label{subsec:rgtg}

In a separate line of work, \citet{khanov2023alignment} proposed reward-guided text generation (RGTG) techniques that do not require any LLM fine-tuning, but can obtain sequences \(\rvy\) with high reward.
This is done by freezing the reference LLM \(\piref\) and at decoding time, the next-token probability \(\piref(y^{i} \mid \rvx, \rvy^{1:i-1})\) is adjusted by a reward model \(r_\phi\).
More specifically, possible values for \(y^{i}\) are scored by a weighted combination of logits of \(\piref\) and the rewards:
\begin{equation*}
  \label{eq:score}
  \textstyle
  \score(y^{i} | \rvx, \rvy^{1:i-1}) = \log \piref(y^{i} \vert \rvx, \rvy^{1:i-1}) + \beta r_\phi(\rvy^{1:i} \vert \rvx) .
\end{equation*}
The next value for \(y^{i}\) is then selected greedily by maximizing their score or by sampling from a softmax distribution of the scores that has a similar form to the RLHF policy in \eqref{eq:policy}:
\begin{equation*}
  % \textstyle
  \softmax(\score(y^{i} \vert \rvx, \rvy^{1:i-1})) = \frac{1}{Z(\rvx, \rvy^{1:i-1})} \piref(y^{i} \vert \rvx, \rvy^{1:i-1}) \exp(\beta r_\phi(\rvy^{1:i} \vert \rvx)) ,
\end{equation*}
where the partition function \(Z(\rvx, \rvy^{1:i-1})\) is now tractable since the summation is now over all possible values of just a \emph{single} variable \(y^{i}\)---it is a summation over possible tokens in the vocabulary.

However, it is unclear whether the resulting distribution is equivalent/approximating the RLHF policy in \eqref{eq:policy}.
\citet{khanov2023alignment} do train the reward model with the BT loss, but it is trained only with complete sequences, i.e.\ \(r_\phi(\rvy \vert \rvx)\), while it is used to score partial sequences, i.e.\ \(r_\phi(\rvy^{1:i} \vert \rvx)\).
Hence, it is unclear whether the inferred scores for partial sequences are reasonable. In \cref{sec:method} we show that reward models trained only with complete sequences can assign arbitrary scores to partial sequences, and in \cref{sec:experiments} we show empirically that the resulting RGTG policy therefore underperforms that of PPO or DPO.
Meanwhile, \citet{deng2023reward} learn the reward model by minimizing a cumulative squared loss to distill a full sequence reward model instead of using the BT loss \eqref{eq:bradley-terry}, making the connection to RLHF policy looser.
\citet{mudgalcontrolled} also distill a partial reward model from a full sequence reward model, but the tokenwise policy is not the marginal of the full-sequence policy.
Nevertheless, tokens are sampled from a different tokenwise RL formulation that follows a similar derivation as RLHF.

\citet{zhao2024probabilistic} proposed to match each of the marginal distribution of \(\pi_\theta(\rvy^{1:i} \vert \rvx)\) by learning a series of parametric functions \(\smash{\{ \psi_{\phi_i} \}_{i=1}^{\abs{\rvy}}}\).
This in turn induces a policy:
\begin{equation*}
  \label{eq:twisted_smc}
  \pi(y^{i} \vert \rvx, \rvy^{1:i-1}) = \frac{1}{Z(\rvx, \rvy^{1:i-1})} \piref(y^{i} \vert \rvx, \rvy^{1:i-1}) \exp(\psi_{\phi_{i}}(\rvy^{1:i} \vert \rvx)) .
\end{equation*}
The generated sequences \(\rvy\) are then approximately equal to the sequences generated by the RLHF policy \eqref{eq:policy}.
However, their method is general and does not specifically target RGTG---indeed, \citet{zhao2024probabilistic} focused on using the implied approximation of the partition function \(Z(\rvx)\). Finally, \citet{rafailov2024qfunction} modifies DPO to obtain a partial-sequence reward model.
\begin{equation*}
  r(\rvy^{1:i} \vert \rvx) = \frac{1}{\beta} \log \frac{\pi_\theta(y^{i} \vert \rvx, \rvy^{1:i-1})}{\piref(y^{i} \vert \rvx, \rvy^{1:i-1})} .
\end{equation*}
Similar to the sequence-based DPO, this reward model is then used to obtain a per-token loss function to fine-tune the LLM and thus, while defining a partial-sequence reward model, is not a RGTG method.

\section{Pitfalls of RGTG and How to Fix Them}
\label{sec:method}

First, we start by analyzing the partial sequence rewards inferred from a reward model trained with full sequences only.
Proof in \cref{app:proof}.

\vspace{0.5em}

\begin{restatable}{theorem}{thmone}
  \label{thm:full_for_partial}
  A reward model $r$ trained to minimize the BT loss \eqref{eq:bradley-terry} on full sequences $\rvy^{1:\abs{\rvy}}$ may assign arbitrary rewards to partial sequences $\rvy^{1:i}$ (where $i< \abs{\rvy}$).
  More precisely, $r(\rvy^{1:i} \vert \rvx) = v_{\rvx,\rvy^{1:i}}$ where $v_{\rvx,\rvy^{1:i}}\in \R$ can be any value.
\end{restatable}

% \vspace{0.25em}

\begin{figure}
  \begin{center}
    % \hspace{-0.25cm}
    % \tikzexternaldisable
    \scalebox{0.8}{
      \includegraphics{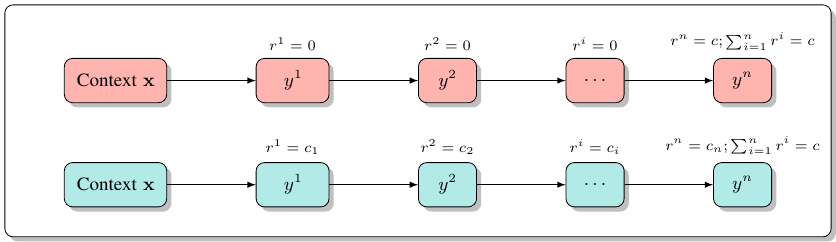}
    }
    % \tikzexternalenable
  \end{center}

  \caption{
    A pathology of using a reward model trained on full-sequence to predict partial sequences in decoding-time RGTG.
    We denote \(r^i = r(y^i \vert \rvx, \rvy^{1:i-1})\).
    While the total reward over the full sequence \(\rvy = (y^1, \dots, y^n)\) might be nonzero \(c\), it could be in the extreme case that the values over previous partial sequences are all zero---this is a perfectly valid result for a sequence-level reward model \textbf{(top)}.
    This means we can have an \emph{unguided} decoding in a \emph{reward-guided} decoding.
    By explicitly training \(r\) on partial sequences, we could avoid this issue \textbf{(bottom)}: While \(\rvy\) might achieve the same final reward \(c\), nonvanishing reward signals over partial sequences could be avoided.
  }
  \label{fig:one}
  \vspace{-1em}
\end{figure}

This leads to an unidentifiability problem---see \cref{fig:one} for an example.
If we learn a reward model based on preferences over full sequences only as proposed by \citet{khanov2023alignment} and \citet{deng2023reward}, then we may not obtain adequate rewards for partial sequences.
As a concrete example, suppose that \(r\) is a reward model such that (\cref{fig:one})
\begin{equation*}
  r(y^i|\rvx,\rvy^{1:i-1})=
  \begin{cases}
    r(\rvy|\rvx) & i = |\rvy|  \\
    0            & i< |\rvy| .
  \end{cases}
\end{equation*}
This reward model satisfies the identity in \eqref{eq:identity} and therefore, could be the solution when minimizing the BT loss \eqref{eq:bradley-terry}.
If we use this reward model to sample from the induced RLHF optimal policy in \eqref{eq:policy}, then the token-level sampling distribution is the same as for the reference LLM \(\piref(y^{i} \mid \rvx, \rvy^{1:i-1})\) for all tokens except the last one.
This is problematic since RLHF generally changes the token-level distribution at each position, not just the last token.
Hence the ARGS method \citep{khanov2023alignment} may utilize a reward model that does not score partial sequences properly, and negatively impact token-by-token generation.

\begin{figure}
  \begin{center}
    \hspace{-0.2cm}
    % \tikzexternaldisable
     \scalebox{0.8}{
    \includegraphics{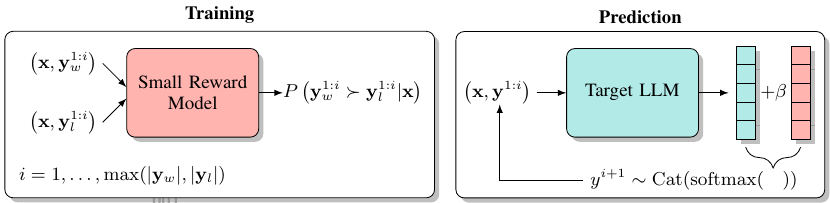}
    }
    % \tikzexternalenable
  \end{center}

  \caption{
    The proposed approach to alleviating the problem in \cref{thm:full_for_partial}.
    First, we (i) train the reward model \(r\) on partial sequences explicitly (when \(\abs{\rvy_w} \neq \abs{\rvy_l}\), we pad to the longest sequence) and (ii) sample from the weighted sum of the logits and the rewards of the next token during decoding.
    This is in contrast to some previous RGTG methods where the reward model is trained on full sequences, but decoding relies on partial sequence scoring. 
  }
  \label{fig:method}
  \vspace{-1em}
\end{figure}

To alleviate this issue, we propose to explicitly train the reward model with partial sequences---still using the BT model in contrast to \citet{deng2023reward}---as follows (\cref{fig:method}).
We create a separate loss function for all prefix lengths $i$:
\begin{equation}
  \textstyle \mathcal{L}_R^i = - \E_{\rvx, \rvy_w, \rvy_l \sim \D} \log \sigma ( r_{\phi} (\rvy_w^{1:i}|\rvx) - r_{\phi} (\rvy_l^{1:i}|\rvx)) . \label{eq:partial-seq-objectives}
\end{equation}

Then, given that full sequence $\rvy_w$ is preferred to full sequence $\rvy_l$, we assume that the partial sequence $\rvy_w^{1:i}$ is also preferred to the partial sequence $\rvy_l^{1:i}$.
Strictly speaking, it is hard for human annotators to compare partial sequences due to their incomplete nature, and most preference datasets do not include preferences over partial sequences.
Nevertheless, we can interpret $\rvy_w^{1:i}$ as the prefix of a winning sequence that is preferred over a losing sequence with prefix $\rvy_l^{1:i}$.  The following lemma shows that the resulting reward model ensures that the probability that
a first partial sequence is preferred to a second partial sequence corresponds to the probability that the first sequence is extended to a winning full sequence while the second sequence is extended to a losing full sequence according to the preference data distribution \(P_\text{data}\). Proof in \cref{app:proof}.

\vspace{0.5em}

\begin{restatable}{lemma}{lemmatwo}
  In the limit of infinite preference data, optimizing a sufficiently expressive reward model according to \eqref{eq:partial-seq-objectives} under the assumption that partial sequences inherit the winning/losing label of full sequences yields a reward model $r_\phi$ with the following property:
  \begin{equation}
    \sigma(r_\phi(\rvy_1^{1:i}|\rvx) - r_\phi(\rvy_2^{1:j}|\rvx)) = P_\textnormal{data}(\rvy_1 \succcurlyeq \rvy_2 | \rvx, \rvy_1^{1:i}, \rvy_2^{1:j}) ,
  \end{equation}
  where \(P_\text{data}\) is the distribution the preference dataset was sampled from and $\rvy_1 \succcurlyeq \rvy_2$ indicates that $\rvy_1$ is preferred to $\rvy_2$.
\end{restatable}

% \vspace{0.5em}

Hence, optimizing the partial-sequence objective \eqref{eq:partial-seq-objectives} for all lengths $i$ determines a reward model for all response prefixes that is adequate in the sense that it induces a distribution over partial sequences that approximates the true underlying preference distribution (due to finite data) instead of assigning arbitrary rewards in the sense of \cref{thm:full_for_partial}.

Once the partial-sequence reward model $r_\phi$ is trained, we can use it to sample the next token $y^i$ conditioned on the previous tokens $\rvx,\rvy^{1:i}$ according to the following conditional distribution:
\begin{equation}
  \pi(y^{i} \vert \rvx,\rvy^{1:i-1}) = \frac{1}{Z(\rvx,\rvy^{1:i-1})} \piref(y^{i} \vert \rvx,\rvy^{1:i-1}) \exp(\beta r_\phi(\rvy^{1:i} \vert \rvx)) .
  \label{eq:token-conditional-distribution}
\end{equation}
%
%This constitutes a tokenwise RGTG method since this distribution.
%\akc{Incomplete sentence?}

Algorithm~\ref{alg:rgtg} summarizes the decoding procedure.
Contrary to the previous approach of \citet{khanov2023alignment}, it directly follows the policy induced by the explicitly trained reward model on partial sequences.
Meanwhile, compared to \citet{deng2023reward}, it uses the standard BT model instead of a custom squared loss function that distills a full-sequence reward model.

Let us now analyze the tokenwise sampling distribution in \eqref{eq:token-conditional-distribution}.
By the definition of conditional distributions, we can rewrite it as a ratio of two partial sequence distributions:
$\pi(y^i \vert \rvx,\rvy^{1:i-1}) = \nicefrac{\pi(\rvy^{1:i} \vert \rvx)}{\pi(\rvy^{1:i-1} \vert \rvx)}$.
However, it is still unclear how this distribution relates to RLHF policies ---- the main point of the tokenwise RGTG methods.
The following theorem shows how the decoding process by following this distribution relates to RLHF-induced policies.
Proof in \cref{app:proof}.

\begin{wrapfigure}[18]{r}{0.5\textwidth}
  \vspace{-2em}

  \begin{minipage}{0.5\textwidth}
    \begin{algorithm}[H]
      \small

      \caption{Decoding with our approach.}
      \label{alg:rgtg}

      \begin{algorithmic}[1]
        \Require Pretrained partial-sequence reward model \(r_\phi\), Prompt $\rvx$, number of candidates $k$, hyperparameter $\beta > 0$, any reference/SFT model $\piref$, generation length $l$
        \Ensure A generated response to $\rvx$ of length $l$
        \For{i = 1 to $l$}
        \State $V^{(k)} = \mathtt{top\_k}(\piref(v \vert \rvx, \rvy^{1:i-1}))$
        \For{$v \in V^{(k)}$}
        \State Reward $r_\phi(\rvy^{1:i-1}, v \vert \rvx))$
        \State Logit $\log \piref(v \vert \rvx, \rvy^{1:i-1})$
        \State $\log \pi(y^i=v \vert \rvx, \rvy^{1:i-1}) =$
        \Statex \qquad\qquad $\log \piref(v \vert \rvx, \rvy^{1:i-1}) + \beta r_\phi(\rvy^{1:i-1}, v \vert \rvx)$
        \EndFor
        \State $y^{i} \sim \mathrm{Cat}(\softmax(\log \pi(y^i \vert \rvx, \rvy^{1:i-1})))$
        \EndFor
        \State \Return  $\rvy^{1:l}$
      \end{algorithmic}
    \end{algorithm}
  \end{minipage}
\end{wrapfigure}

\vspace{0.5em}

\begin{restatable}{theorem}{thmthree}
  \label{thm:ratio}
  Given a reward model trained according to the partial-sequence BT objective in \eqref{eq:partial-seq-objectives}, the induced token generation distribution \(\pi\) \eqref{eq:token-conditional-distribution} is proportional to the ratio:
  \begin{equation} \label{eq:multiple-rlhf}
    \pi(y^i \vert \rvx,\rvy^{1:i-1}) \propto \frac{\pi_{\textnormal{RLHF},i}(\rvy^{1:i} \vert \rvx)}{\pi_{\textnormal{RLHF},i-1}(\rvy^{1:i-1} \vert \rvx)}
  \end{equation}
  where \(\pi_{\textnormal{RLHF},i}\) and \(\pi_{\textnormal{RLHF},i-1}\) are two distinct policies over prefix sequences of length $i$ and $i-1$, respectively, induced by RLHF optimization \eqref{eq:rlhf}.
\end{restatable}

% \vspace{0.5em}

% \subsection{Discussion}

Ideally, we would like a decoding procedure that samples the next token from a distribution that is mathematically equivalent to the conditional distribution resulting from an RLHF over full sequences.
However, as shown in \cref{thm:ratio}, a partial-sequence reward model $r_\phi$ leads to multiple RLHF decoding policies with different conditional distributions for each prefix length $i$.
Hence, it is not possible to have equivalence with a single RLHF policy, e.g.\ as obtained via PPO or DPO.

One may then ask: Which RLHF policy is best?
We argue that none of them is necessarily better than the others since they simply arise from considering different prefix lengths.
Note that the reward model $r_\phi$ leads to a distribution that approximates the true underlying preference distribution on partial sequences.
The problem is inherent to RLHF which takes a reference LLM with a consistent distribution over response prefixes induced by a reward model and yields different decoding policies for different prefix lengths.

Since all the resulting RLHF decoding policies have merit, one could argue that we can keep things simple by selecting only one policy, perhaps the RLHF policy induced by full sequence preferences (i.e., $\pi_{\text{RLHF}}(\rvy|\rvx)$).
However, as discussed by \citet{rafailov2024qfunction} and \citet{zhao2024probabilistic}, a conditional distribution over full sequences does not give us an immediate procedure for token-wise sampling.
Mathematically, we can derive a token-level policy from a full-sequence policy as follows:
\begin{align*}
  \pi_{\text{RLHF}}(y^i \vert \rvx,\rvy^{1:i-1}) = \frac{\pi_{\text{RLHF}}(\rvy^{1:i} \vert \rvx)}{\pi_{\text{RLHF}}(\rvy^{1:i-1} \vert \rvx)} = \frac{\sum_{\rvy^{i+1:|\rvy|}} \pi_{\text{RLHF}}(\rvy|\rvx)}{\sum_{\rvy^{i:|\rvy|}} \pi_{\text{RLHF}}(\rvy|\rvx)} .
\end{align*}
However, the summations in the above equation are exponentially large in the length $|\rvy|$ of the sequences.
This exponential complexity was also noted by \citet{zhao2024probabilistic} who proposed a twisted sequential Monte Carlo technique to approximate the computation and mitigate the exponential complexity.
In contrast, our approach embraces the multitude of RLHF policies and leverages them in a linear time decoding procedure without any approximation of the partial sequence RLHF policies.
The ratio policy \eqref{eq:token-conditional-distribution} described here can thus be seen as a necessary tradeoff if one wants to perform tokenwise RGTG without the pathology in \cref{thm:full_for_partial}.

%\section{Theoretical Analysis}
%\label{sec:analysis}
%\input{contents/04_analysis}

\section{Related Work}
\label{sec:related-work}
% \vspace{-2em}

\textbf{Language model alignment}\quad Simple fine-tuning and instruction tuning~\citep{wei2021finetuned} are ways to align LLMs to labeled data. Recently, RLHF methods~\citep{christiano2017deep,ziegler2019fine,lee2021pebble,nakano2021webgpt,snell2022offline} have provided a direct method to align LLMs to human preferences. 
The PPO algorithm has been especially popular and has shown promising results for a range of tasks~\citep{askell2021general, bai2022training,ouyang2022training}. 
However, training RL models is compute intensive and researchers have turned their attention to supervised fine-tuning methods that can learn directly from preference data. 
\citet{liu2023chain} turns the preference data into prompts with which they fine-tune the LLM. 
\citet{dong2023raft} uses the reward model to filter the training set to better fine-tune the model. 
DPO \citep{rafailov2023direct,rafailov2024qfunction} models the LLM itself as a Bradley-Terry model and optimizes the RLHF objective without any need for RL. TDPO~\citep{zengtoken} incorporates token-level KL divergence into the DPO objective to improve content diversity.
These methods, however,  fine-tune the base LLM, which can be expensive as we scale. Some works have attempted to improve alignment by gathering more fine-grained rewards by using either LLMs~\citep{cao2024sparserewardsenhancingreinforcement} or human annotators~\citep{wu2024fine}.

\textbf{Guided decoding}\quad There has been prior work in guided decoding using sequence-level~\citep{welleck2022naturalprover,uesato2022solving, lightman2023let, krishna2022rankgen,li2023making, khalifa2023grace, yao2023tree} and token-level value functions \citep{dathathri2019plug, krause2021gedi, yang2021fudge,chaffin2022ppl, liu2023attribute}. PPLM ~\citep{dathathri2019plug}  uses the gradients from an attribute classifier to guide LLM generation. Gedi~\citep{krause2021gedi} uses attribute conditioned language models as discriminators to update LLM generation probabilities using Bayes rule. These algorithms are different from our work as they do not align LLMs using human preference data. \citet{deng2023reward} use a reward model trained on preference data in the decoding process, however, they use a cumulative squared loss function that is different from the RLHF framework. 
\citet{mudgalcontrolled} uses a similar loss function with the key difference that instead of training with samples from a preference dataset, they take as input a full sequence reward model and train a partial sequence value function based on roll-outs (i.e., sampled token sequences) from the base LLM. 
Therefore, for each new base LLM, the value function needs to be retrained with new roll-outs, limiting portability to new or updated language models. 
The closest work to our method is \citet{khanov2023alignment}, which is also based on the Bradley-Terry model, but they use a reward model trained on full sequences, which we have argued can lead to pitfalls.
Different from our work, \citet{zhao2024probabilistic} present a reward-guided decoding method based on sequential Monte Carlo and show that it can approximate RLHF.

\textbf{Partial Rewards} \quad Outside of preference data alignment and RLHF, prior work in reinforcement learning for language modeling has looked at partial reward models for improving text generation. \cite{hao2022teacher} show that a sequence to sequence model trained with supervised learning is a valid partial reward model for text generation under a Markov decision process. ~\cite{lee2023learning} do not train an explicit reward model but instead introduce a ranking function which can rank the next token for partial sequences. Both these methods modify language model training.

\section{Experiments}
\label{sec:experiments}

\begin{table*}[t]
  \centering
  \footnotesize
  \begin{tabular}[t]{cccc}
    \toprule
    \multicolumn{4}{c}{\textbf{TL;DR Summarization}}                                    \\
    \midrule
    \textbf{Method} & \textbf{LLM} & \textbf{Single \(\rvy\)?} & $r \pm \text{SE}$      \\
    \midrule
    Top-\(k\)       & frozen       & yes                       & -0.11$\pm$0.28         \\
    CD              & frozen       & yes                       & 0.32$\pm$0.33          \\
    ARGS            & frozen       & yes                       & 1.57$\pm$0.21          \\
    \emph{PARGS-G}  & frozen       & yes                       & 2.06$\pm$0.20          \\
    \emph{PARGS}    & frozen       & yes                       & \textbf{2.36$\pm$0.20} \\
    \midrule
    Best-of-\(N\)   & frozen       & no                        & 2.2 $\pm$0.19          \\
    DPO             & trained      & yes                       & 0.81$\pm$0.26          \\
    PPO             & trained      & yes                       & 2.41$\pm$0.23          \\
    \bottomrule
  \end{tabular}
  \hfill
  % \hspace{0cm}
  %
  \begin{tabular}[t]{cccc}
    \toprule
    \multicolumn{4}{c}{\textbf{HH Dialogue}}                                             \\
    \midrule
    \textbf{Method} & \textbf{LLM} & \textbf{Single \(\rvy\)?} & $r \pm \text{SE}$       \\
    \midrule
    Top-\(k\)       & frozen       & yes                       & -1.42$\pm$0.21          \\
    CD              & frozen       & yes                       & -1.08$\pm$0.21          \\
    ARGS            & frozen       & yes                       & -0.97$\pm$0.19          \\
    \emph{PARGS-G}  & frozen       & yes                       & -0.97$\pm$0.18          \\
    \emph{PARGS}    & frozen       & yes                       & \textbf{-0.88$\pm$0.19} \\
    \midrule
    Best-of-\(N\)   & frozen       & no                        & 0.17 $\pm$0.18          \\
    DPO             & trained      & yes                       & -0.79$\pm$0.31          \\
    \bottomrule
  \end{tabular}

  \caption{Average reward (over 100 samples) $\pm$ standard error for the TL;DR summarization and HH dialogue tasks.  The best technique that freezes the LLM and generates a single response $\rvy$ is bolded.}
  \label{tab:reward}
  % \vspace{-0.5em}
\end{table*}

\begin{table*}[t]
  \centering
  \footnotesize
  \begin{tabular}[t]{cccc}
    \toprule
    \multicolumn{4}{c}{\textbf{TL;DR Summarization}}                              \\
    \midrule
    \textbf{Method A} & \textbf{vs} & \textbf{Method B} & \textbf{Win-Tie ($\%$)} \\
    \midrule
    PARGS             &             & CD                & 75 - 0                  \\
    PARGS             &             & ARGS              & 73 - 0                  \\
    PARGS             &             & Best-of-\(N\)     & 55 - 0                  \\
    PARGS             &             & DPO               & 59 - 1                  \\
    PARGS             &             & PPO               & 56 - 0                  \\
    \bottomrule
  \end{tabular}
  \hfill
  \begin{tabular}[t]{cccc}
    \toprule
    \multicolumn{4}{c}{\textbf{HH Dialogue}}                                      \\
    \midrule
    \textbf{Method A} & \textbf{vs} & \textbf{Method B} & \textbf{Win-Tie ($\%$)} \\
    \midrule
    PARGS             &             & CD                & 52 - 8
    \\
    PARGS             &             & ARGS              & 49 - 11
    \\
    PARGS             &             & Best-of-\(N\)     & 36 - 11                 \\
    PARGS             &             & Top-\(k\)         & 56  - 15                \\
    PARGS             &             & DPO               & 27  - 14                \\
    \bottomrule
  \end{tabular}

  \caption{GPT-4 evaluation based on the win-tie rate of PARGS over different baselines on TL;DR summarization with GPT$2$-large, and on HH dialogue generation with Llama-2-7b.}
  \label{tab:gpt4}
  % \vspace{-0.5em}
\end{table*}

We evaluate our proposed approach, which we call \textbf{P}artial \textbf{A}lignment as \textbf{R}eward-\textbf{G}uided \textbf{S}ampling (PARGS)---in contrast to ARGS which considers full sequences and greedy decoding instead of sampling---on two generation tasks: summarization and dialogue generation.
%We assume access to a preference dataset with labels for accepted and rejected answers.

\subsection{Setup}
\label{subsec:setup}
\textbf{Summarization task}\quad We use the Reddit TL;DR dataset~\citep{volske2017tl}, where, the context $\vx$ is a post on the Reddit forum and $\vy$ is the summary of the post.
We use the human preference dataset from \citet{stiennon2020learning} to train the reward model and the relevant baselines.
Our base summarization model is GPT2-large, fine-tuned on the TL;DR training set.
We use a pre-trained reward model based on the DeBerta-v3-large architecture
% \footnote{\href{https://huggingface.co/OpenAssistant/reward-model-deberta-v3-large-v2}{OpenAssistant/reward-model-deberta-v3-large-v2}} 
and train it with partial sequences for an additional epoch.
Our baselines include top-\(k\) sampling~\citep{fan2018hierarchical}, Best-of-\(N\) generation, which involves sampling \(N\) sequences from  reference LLM (\(N=10\) for all our experiments) and returning the best one according to the reward model, RLHF models based on PPO and DPO, the reward-base decoding method ARGS~\citep{khanov2023alignment} and controlled decoding \citep[CD;][]{mudgalcontrolled}. We use CD-Fudge as the baseline in all our CD experiments, noting that its performance is similar to CD-Q (see Table 4 in \citet{mudgalcontrolled}).

\textbf{Dialogue task}\quad
Next, we evaluate our model on single-turn dialogue using the Anthropic Helpful and Harmless \citep[HH;][]{bai2022training} dataset.
The goal is to generate a helpful and harmless response to a general purpose query.
Each sample provides a prompt $\rvx$ and two responses $\rvy$ with a label indicating the preferred response.
% Here $\rvx$ is the human dialogue and $\rvy$ is the response from the assistant.
We use Llama-2-7b as the base model and DeBerta-v3 as the reward model which is about $20\times$ smaller.
Details in \ref{app:training}.

\textbf{Fine-grained text generation task}\quad
We also evaluate our model on the UltraFeedback dataset \citep{cui2024ultrafeedback}.
% The goal is to generate sequences that satisfy the following four aspects: instruction-following, truthfulness, honesty and helpfulness.
We use Zephyr-7B as the base LLM and Phi-1.5 (1.3 billion parameters) as the reward model.

\textbf{Machine translation task}\quad
We perform additional experiments on machine-translation on the IWSLT-2017 dataset \citep{cettolo-etal-2017-overview}.
We used the post-edit dataset from \citet{kreutzer2020correct} on the IWSLT-2017 English-German dataset to provide token-wise reward signals.
% For this task we only compare out methods against ARGS and greedy decoding. 
We use Gemma-2b as both the base model and the reward model.
The evaluation is based on the standard BLUE score.

\textbf{Evaluation}\quad
Following \citet{khanov2023alignment}, we compare all methods based on \emph{average reward}, on the test samples, as measured by the reward model.
% A higher reward indicates better alignment with the reward model.
We use a \emph{different} full-sequence reward model and not the partial-sequence reward model (that we trained for our algorithm) to evaluate the models.
Since evaluating language generation, especially unconditionally, is nuanced and human evaluation is very expensive, we use GPT-4-based evaluation, which has been shown to align with human assessment~\citep{zheng2023judging,rafailov2023direct}.
% We adopt GPT-4 based evaluation as a proxy of human evaluation. 
Following \citet{chiang2023vicuna} we construct prompts for the two tasks and ask GPT-4 to score and rank response pairs.
We randomly shuffle the order of the responses to mitigate position bias~\citep{zheng2023judging}. Finally, we use the Rouge-L score \citep{lin2004rouge} and the BLEU score to evaluate the dialogue and translation tasks, respectively.

\subsection{Results}
\label{sec:results}

\Cref{tab:reward} (left) shows the average reward for the summaries generated by the different algorithms as measured by the reward model. PARGS achieves the best average reward among the techniques that keep the LLM frozen and generate a single response $\rvy$.  We also note that PARGS outperforms DPO and is competitive with PPO based RLHF that incurs a large cost to fine-tune the LLM, and Best-of-N that incurs significant overhead to generate multiple responses. Upon siginficance testing we observed PARGS to be significantly better than all algorithms expect PPO. Details in \cref{app:significance}. Note that we also evaluate our algorithm with greedy decoding (PARGS-G) for a direct comparison with ARGS.

Similarly, \cref{tab:reward} (right) presents average rewards for the responses of the different algorithms on the HH dialogue task. Note that in this setting, the reward model is $20\times$ smaller than the base LLM.  Again, PARGS achieved the highest reward among the techniques that freeze the LLM and generate a single response.
We observe that Best-of-\(N\) achieved the highest average reward followed by DPO, but incurred overhead to generate multiple responses and fine-tune the LLM  respectively.
Finally, \cref{tab:UF reward} (left) presents average rewards on the UltraFeedback dataset. We observe that PARGS outperforms all methods except Best-of-\(N\). Significance testing (see \cref{app:significance}) reveals that PARGS is significantly better.

% \begin{figure*}[t]
% 	\centering
% 	\includegraphics[width=\linewidth]{figures/tldr3.pdf}

% 	\caption{
% 		Summary statistics of rewards attained by different algorithms on TLDR summarization.
% 	}
% 	\label{fig:tldr}
% \end{figure*}

\begin{table*}[t]

  \centering
  \footnotesize
  \begin{tabular}[t]{cccc}
    \toprule
    \multicolumn{4}{c}{\textbf{Ultra Feedback}}                                         \\
    \midrule
    \textbf{Method} & \textbf{LLM} & \textbf{Single \(\rvy\)?} & $r \pm \text{SE}$      \\
    \midrule
    Top-\(k\)       & frozen       & yes                       & -0.18$\pm$0.12         \\
    CD              & frozen       & yes                       & -0.04$\pm$0.01         \\
    ARGS            & frozen       & yes                       & 0.01$\pm$0.12          \\
    \emph{PARGS}    & frozen       & yes                       & \textbf{0.21$\pm$0.09} \\
    \emph{PARGS-G}  & frozen       & yes                       & \textbf{0.21$\pm$0.12} \\
    \midrule
    DPO             & trained      & yes                       & -0.57$\pm$0.09         \\
    Best-of-\(N\)   & frozen       & no                        & 1.15 $\pm$0.08         \\
    \bottomrule
  \end{tabular}
  \begin{tabular}[t]{cccc}
    \toprule
    \multicolumn{4}{c}{\textbf{Ultra Feedback}}                                   \\
    \midrule
    \textbf{Method A} & \textbf{vs} & \textbf{Method B} & \textbf{Win-Tie ($\%$)} \\
    \midrule
    PARGS             &             & CD                & 53 - 13                 \\
    PARGS             &             & ARGS              & 42 - 23                 \\
    PARGS             &             & Best-of-\(N\)     & 29 - 19                 \\
    PARGS             &             & Top-\(k\)         & 52 - 15                 \\

    PARGS             &             & DPO               & 65 - 7                  \\
    \bottomrule
  \end{tabular}

  \caption{Average reward (100 samples) $\pm$ std.\ error and GPT-4 evaluation for Ultra Feedback.}
  \label{tab:UF reward}
  \vspace{-1em}
\end{table*}

Next we evaluate PARGS using GPT-4. The prompt used to probe GPT-4 is presented in Appendix~\ref{app:gpt-4}.
\Cref{tab:gpt4} reports the win-tie rate (i.e., percentage of utterances where GPT-4 finds PARGS' response to be better than or equivalent to those of the baselines).  Table~\ref{tab:gpt4} (left) shows that PARGS has a higher win-tie rate compared to all the methods, especially ARGS, for TL;DR summarization.
As noted by others~\cite{rafailov2023direct}, Best-of-\(N\) is a strong baseline, but it is computationally intensive.  On HH, we observe (\cref{tab:gpt4} right) that PARGS is better than CD and ARGS, but worse than Best-of-\(N\) and DPO.
As we scale training based alignment methods, e.g., DPO become prohibitive. On UltraFeedback (\cref{tab:UF reward} right) we observe that PARGS outperforms all methods except  Best-of-\(N\). We perform human evaluation for PARGS against ARGS, CD and DPO on UltraFeedback. PARGS wins against all three baselines. Detailed results in \cref{app:humanEval}.

% \begin{wraptable}[4]{r}{0.3\textwidth}
%   \vspace{-2em}
%   \centering
%   \footnotesize

%   \caption{BLEU Score on IWSLT-17 English to German}
%   \label{tab:MT}

%   \begin{tabular}{ccc}
%     \toprule
%     \textbf{Method} & \textbf{BLEU \(\uparrow\)} \\
%     \midrule
%     Greedy          & 31.7  $\pm$  3.6           \\
%     ARGS            & 29.4  $\pm$  3.4           \\
%     PARGS-G         & 33.2  $\pm$  3.5           \\
%     \bottomrule
%   \end{tabular}
%   % \vspace{1em}
% \end{wraptable}

We perform additional experiments on machine-translation on the IWSLT-2017 dataset \citep{cettolo-etal-2017-overview}.
We used the post-edit dataset from \citet{kreutzer2020correct} on the IWSLT-2017 English-German dataset to provide token-wise reward signals. We use Gemma-2b as both the base model and the reward model. The evaluation is based on the standard BLUE score.

\begin{wraptable}[11]{r}{0.3\textwidth}
  % \vspace{-2em}
  \centering
  \footnotesize

  \caption{BLEU Score on IWSLT-17 English-German}
  \label{tab:MT}

  \begin{tabular}{ccc}
    \toprule
    \textbf{Method} & \textbf{BLEU \(\uparrow\)} \\
    \midrule
    Greedy          & 31.7  $\pm$  3.6           \\
    ARGS            & 29.4  $\pm$  3.4           \\
    PARGS-G         & 33.2  $\pm$  3.5           \\
    \bottomrule
  \end{tabular}
  % \vspace{1em}
\end{wraptable}

The translation direction is English to German and the edited/corrected sequence is considered the winning sequence. \cref{tab:MT} compares ARGS, PARGS-G and greedy decoding. We observe that applying ARGS reduces the BLUE score of the greedy baseline where PARGS-G increases it by 1.5 on average.

\begin{wraptable}[6]{r}{0.3\textwidth}
  \vspace{-4em}
  \centering
  \footnotesize

  \caption{Diversity Results}
  \label{tab:diversity}

  \begin{tabular}{ccc}
    \toprule
    \textbf{Method} & \textbf{ROUGE-L \(\downarrow\)} \\
    \midrule
    Top-\(k\)       & 0.230 $\pm 0.011$               \\
    DPO             & 0.206 $\pm 0.006$               \\
    PARGS           & 0.203 $\pm 0.008$               \\
    \bottomrule
  \end{tabular}
\end{wraptable}

% DPO and Best-of-$N$ also trade diversity for accuracy as discussed next.
%On the other hand PARGS pays a small decoding overhead and can align the LLM to human preferences well.

% \begin{figure*}[t]
% 	\centering
% 	\includegraphics[width=\linewidth]{figures/hh.pdf}

% 	\caption{
%     Summary statistics of rewards attained by different algorithms on HH dialogue generation.
% 	}
% 	\label{fig:hh}
% \end{figure*}

% \begin{table*}[t]
% 	\centering
% 	\begin{tabular}{cccc}
% 		\toprule
% 		Method A & vs & Method B  & Win-Tie ($\%$) \\
% 		\midrule
% 		PARGS    &    & Top-K     & 73             \\
% 		PARGS    &    & ARGS      & 60             \\
% 		PARGS    &    & Best-of-N & 47             \\
% 		PARGS    &    & DPO       & 41             \\
% 		\bottomrule
% 	\end{tabular}
% 	\caption{GPT-4 evaluation based on the win-rate of our algorithm vs different baselines on Anthropic-HH dialogue with Llama2-7b. Note that the reward model used is trained on the much smaller DeBerta-v3-large architecture. }
% 	\label{tab:hh-gpt}
% \end{table*}
%

We evaluate the diversity of generation on 50 samples from the UltraFeedback dataset. We compare different methods by generating \(10\) responses for each prompt, evaluating the Rouge-L score between each generated pair and averaging it over all the samples.
A lower Rouge-L score indicates a greater diversity.
\cref{tab:diversity} shows that PARGS generates the most diverse responses compared to top-\(K\) and DPO.

Note that Best-of-\(N\) generates $N\times$ the number of samples using top-\(K\) generation.

\textbf{Additional Results}  A discussion on decoding costs is presented in Appendix~\ref{app:cost}. A study on effect of the hyper-parameter $\beta$ and K in top-K is presented in \cref{app:sensitivity}. We present an empirical analysis validating our assumption that the partial sequence of a winning sequence wins over the partial sequence of a losing sequence in~\cref{app:analysis}.

\section{Conclusion}
\label{sec:conclusion}

We have discussed the pitfalls in tokenwise, decoding-time reward-guided text generation (RGTG) with reward models trained on full sequences. 
These pitfalls can lead to inadequate reward during the autoregressive decoding process and may lead to subpar performance.
To alleviate this, we train reward models on partial sequences and then sample from the implied per-token text generation policy during decoding.
We proved that this policy is a ratio of \emph{two} distinct reinforcement learning from human feedback (RLHF) policies. 
This means that this policy is not equivalent to the standard offline RLHF methods. 
However, we have also shown that it is intractable to obtain a tokenwise policy that is equivalent to a \emph{single} RLHF policy. 
Our experiment results validated our approach: it performs better than  recent RGTG methods such as ARGS, that leverages full-sequence reward models, and CD. We discuss limitations of our work in Appendix~\ref{app:lim}.

\section*{Acknowledgments and Disclosure of Funding}
Resources used in this work were provided by the Province of Ontario, the Government of Canada through CIFAR, companies sponsoring the Vector Institute \url{https://vectorinstitute.ai/partners/} and the Natural Sciences and Engineering Council of Canada. AR thanks Apple for support through the Waterloo Apple PhD Fellowship, Natural Sciences and Engineering Council of Canada for its support through the CGS-D program, and the David R. Cheriton Graduate Scholarship. JG thanks Microsoft Research for support through its PhD Scholarship Programme and the International Max Planck Research School for Intelligent Systems (IMPRS-IS). AK thanks Rob Brekelmans for a fruitful discussion.

% \section*{Ethics Statement}
% Authors can add an optional ethics statement to the paper. 
% For papers that touch on ethical issues, this section will be evaluated as part of the review process. The ethics statement should come at the end of the paper. It does not count toward the page limit, but should not be more than 1 page. 

\bibliography{colm2025_conference}

\begin{thebibliography}{53}
\providecommand{\natexlab}[1]{#1}
\providecommand{\url}[1]{\texttt{#1}}
\expandafter\ifx\csname urlstyle\endcsname\relax
  \providecommand{\doi}[1]{doi: #1}\else
  \providecommand{\doi}{doi: \begingroup \urlstyle{rm}\Url}\fi

\bibitem[Askell et~al.(2021)Askell, Bai, Chen, Drain, Ganguli, Henighan, Jones, Joseph, Mann, DasSarma, et~al.]{askell2021general}
Amanda Askell, Yuntao Bai, Anna Chen, Dawn Drain, Deep Ganguli, Tom Henighan, Andy Jones, Nicholas Joseph, Ben Mann, Nova DasSarma, et~al.
\newblock A general language assistant as a laboratory for alignment.
\newblock \emph{arXiv preprint arXiv:2112.00861}, 2021.

\bibitem[Bai et~al.(2022)Bai, Jones, Ndousse, Askell, Chen, DasSarma, Drain, Fort, Ganguli, Henighan, et~al.]{bai2022training}
Yuntao Bai, Andy Jones, Kamal Ndousse, Amanda Askell, Anna Chen, Nova DasSarma, Dawn Drain, Stanislav Fort, Deep Ganguli, Tom Henighan, et~al.
\newblock Training a helpful and harmless assistant with reinforcement learning from human feedback.
\newblock \emph{arXiv preprint arXiv:2204.05862}, 2022.

\bibitem[Bradley \& Terry(1952)Bradley and Terry]{bradleyterry1952paired}
Ralph~Allan Bradley and Milton~E Terry.
\newblock Rank analysis of incomplete block designs: The method of paired comparisons.
\newblock \emph{Biometrika}, 39\penalty0 (3/4), 1952.

\bibitem[Brown et~al.(2020)Brown, Mann, Ryder, Subbiah, Kaplan, Dhariwal, Neelakantan, Shyam, Sastry, Askell, et~al.]{brown2020gpt3}
Tom Brown, Benjamin Mann, Nick Ryder, Melanie Subbiah, Jared~D Kaplan, Prafulla Dhariwal, Arvind Neelakantan, Pranav Shyam, Girish Sastry, Amanda Askell, et~al.
\newblock Language models are few-shot learners.
\newblock In \emph{NeurIPS}, 2020.

\bibitem[Cao et~al.(2024)Cao, Shu, Yu, Zhu, Wichers, Liu, and Meng]{cao2024sparserewardsenhancingreinforcement}
Meng Cao, Lei Shu, Lei Yu, Yun Zhu, Nevan Wichers, Yinxiao Liu, and Lei Meng.
\newblock Beyond sparse rewards: Enhancing reinforcement learning with language model critique in text generation.
\newblock \emph{arXiv preprint arXiv:2401.07382}, 2024.

\bibitem[Cettolo et~al.(2017)Cettolo, Federico, Bentivogli, Niehues, St{\"u}ker, Sudoh, Yoshino, and Federmann]{cettolo-etal-2017-overview}
Mauro Cettolo, Marcello Federico, Luisa Bentivogli, Jan Niehues, Sebastian St{\"u}ker, Katsuhito Sudoh, Koichiro Yoshino, and Christian Federmann.
\newblock Overview of the {IWSLT} 2017 evaluation campaign.
\newblock In \emph{Proceedings of the 14th International Conference on Spoken Language Translation}, pp.\  2--14, Tokyo, Japan, December 14-15 2017. International Workshop on Spoken Language Translation.
\newblock URL \url{https://aclanthology.org/2017.iwslt-1.1}.

\bibitem[Chaffin et~al.(2022)Chaffin, Claveau, and Kijak]{chaffin2022ppl}
Antoine Chaffin, Vincent Claveau, and Ewa Kijak.
\newblock {PPL-MCTS}: Constrained textual generation through discriminator-guided {MCTS} decoding.
\newblock In \emph{NAACL}, 2022.

\bibitem[Chiang et~al.(2023)Chiang, Li, Lin, Sheng, Wu, Zhang, Zheng, Zhuang, Zhuang, Gonzalez, Stoica, and Xing]{chiang2023vicuna}
Wei-Lin Chiang, Zhuohan Li, Zi~Lin, Ying Sheng, Zhanghao Wu, Hao Zhang, Lianmin Zheng, Siyuan Zhuang, Yonghao Zhuang, Joseph~E. Gonzalez, Ion Stoica, and Eric~P. Xing.
\newblock {Vicuna}: An open-source chatbot impressing {GPT-4} with 90\%* {ChatGPT} quality, March 2023.
\newblock URL \url{https://lmsys.org/blog/2023-03-30-vicuna/}.

\bibitem[Christiano et~al.(2017)Christiano, Leike, Brown, Martic, Legg, and Amodei]{christiano2017deep}
Paul~F Christiano, Jan Leike, Tom Brown, Miljan Martic, Shane Legg, and Dario Amodei.
\newblock Deep reinforcement learning from human preferences.
\newblock In \emph{NIPS}, 2017.

\bibitem[Dathathri et~al.(2019)Dathathri, Madotto, Lan, Hung, Frank, Molino, Yosinski, and Liu]{dathathri2019plug}
Sumanth Dathathri, Andrea Madotto, Janice Lan, Jane Hung, Eric Frank, Piero Molino, Jason Yosinski, and Rosanne Liu.
\newblock Plug and play language models: A simple approach to controlled text generation.
\newblock In \emph{ICLR}, 2019.

\bibitem[Deng \& Raffel(2023)Deng and Raffel]{deng2023reward}
Haikang Deng and Colin Raffel.
\newblock Reward-augmented decoding: Efficient controlled text generation with a unidirectional reward model.
\newblock In \emph{EMNLP}, 2023.

\bibitem[Dong et~al.(2023)Dong, Xiong, Goyal, Zhang, Chow, Pan, Diao, Zhang, KaShun, and Zhang]{dong2023raft}
Hanze Dong, Wei Xiong, Deepanshu Goyal, Yihan Zhang, Winnie Chow, Rui Pan, Shizhe Diao, Jipeng Zhang, SHUM KaShun, and Tong Zhang.
\newblock {RAFT}: Reward ranked finetuning for generative foundation model alignment.
\newblock \emph{TMLR}, 2023.

\bibitem[Fan et~al.(2018)Fan, Lewis, and Dauphin]{fan2018hierarchical}
Angela Fan, Mike Lewis, and Yann Dauphin.
\newblock Hierarchical neural story generation.
\newblock In \emph{ACL}, 2018.

\bibitem[Ganqu~Cui et~al.(2024)Ganqu~Cui, Ding, Yao, He, Zhu, Ni, Xie, Xie, Lin, Liu, and Sun]{cui2024ultrafeedback}
Lifan~Yuan Ganqu~Cui, Ning Ding, Guanming Yao, Bingxiang He, Wei Zhu, Yuan Ni, Guotong Xie, Ruobing Xie, Yankai Lin, Zhiyuan Liu, and Maosong Sun.
\newblock Ultrafeedback: Boosting language models with scaled ai feedback.
\newblock In \emph{ICML}, 2024.

\bibitem[Hao et~al.(2022)Hao, Liu, and Mou]{hao2022teacher}
Yongchang Hao, Yuxin Liu, and Lili Mou.
\newblock Teacher forcing recovers reward functions for text generation.
\newblock \emph{NeurIPS}, 2022.

\bibitem[Kaplan et~al.(2020)Kaplan, McCandlish, Henighan, Brown, Chess, Child, Gray, Radford, Wu, and Amodei]{kaplan2020scaling}
Jared Kaplan, Sam McCandlish, Tom Henighan, Tom~B Brown, Benjamin Chess, Rewon Child, Scott Gray, Alec Radford, Jeffrey Wu, and Dario Amodei.
\newblock Scaling laws for neural language models.
\newblock \emph{arXiv preprint arXiv:2001.08361}, 2020.

\bibitem[Khalifa et~al.(2023)Khalifa, Logeswaran, Lee, Lee, and Wang]{khalifa2023grace}
Muhammad Khalifa, Lajanugen Logeswaran, Moontae Lee, Honglak Lee, and Lu~Wang.
\newblock Grace: Discriminator-guided chain-of-thought reasoning.
\newblock In \emph{EMNLP}, 2023.

\bibitem[Khanov et~al.(2024)Khanov, Burapacheep, and Li]{khanov2023alignment}
Maxim Khanov, Jirayu Burapacheep, and Yixuan Li.
\newblock Alignment as reward-guided search.
\newblock In \emph{ICLR}, 2024.

\bibitem[Krause et~al.(2021)Krause, Gotmare, McCann, Keskar, Joty, Socher, and Rajani]{krause2021gedi}
Ben Krause, Akhilesh~Deepak Gotmare, Bryan McCann, Nitish~Shirish Keskar, Shafiq Joty, Richard Socher, and Nazneen~Fatema Rajani.
\newblock {GeDi}: Generative discriminator guided sequence generation.
\newblock In \emph{EMNLP}, 2021.

\bibitem[Kreutzer et~al.(2020)Kreutzer, Berger, and Riezler]{kreutzer2020correct}
Julia Kreutzer, Nathaniel Berger, and Stefan Riezler.
\newblock Correct me if you can: Learning from error corrections and markings.
\newblock In \emph{Proceedings of the 22nd Annual Conference of the European Association for Machine Translation}, 2020.

\bibitem[Krishna et~al.(2022)Krishna, Chang, Wieting, and Iyyer]{krishna2022rankgen}
Kalpesh Krishna, Yapei Chang, John Wieting, and Mohit Iyyer.
\newblock Rankgen: Improving text generation with large ranking models.
\newblock In \emph{EMNLP}, 2022.

\bibitem[Lee et~al.(2021)Lee, Smith, and Abbeel]{lee2021pebble}
Kimin Lee, Laura~M Smith, and Pieter Abbeel.
\newblock {PEBBLE}: Feedback-efficient interactive reinforcement learning via relabeling experience and unsupervised pre-training.
\newblock In \emph{ICML}, 2021.

\bibitem[Lee et~al.(2023)Lee, Lee, and Hwang]{lee2023learning}
Youngwon Lee, Jinu Lee, and Seung-won Hwang.
\newblock Learning to rank generation with pairwise partial rewards.
\newblock In \emph{EMNLP}, 2023.

\bibitem[Li et~al.(2023)Li, Lin, Zhang, Fu, Chen, Lou, and Chen]{li2023making}
Yifei Li, Zeqi Lin, Shizhuo Zhang, Qiang Fu, Bei Chen, Jian-Guang Lou, and Weizhu Chen.
\newblock Making language models better reasoners with step-aware verifier.
\newblock In \emph{ACL}, 2023.

\bibitem[Lightman et~al.(2023)Lightman, Kosaraju, Burda, Edwards, Baker, Lee, Leike, Schulman, Sutskever, and Cobbe]{lightman2023let}
Hunter Lightman, Vineet Kosaraju, Yuri Burda, Harrison Edwards, Bowen Baker, Teddy Lee, Jan Leike, John Schulman, Ilya Sutskever, and Karl Cobbe.
\newblock Let's verify step by step.
\newblock In \emph{ICLR}, 2023.

\bibitem[Lin(2004)]{lin2004rouge}
Chin-Yew Lin.
\newblock Rouge: A package for automatic evaluation of summaries.
\newblock In \emph{Text summarization branches out}, pp.\  74--81, 2004.

\bibitem[Liu et~al.(2023{\natexlab{a}})Liu, Sferrazza, and Abbeel]{liu2023chain}
Hao Liu, Carmelo Sferrazza, and Pieter Abbeel.
\newblock Chain of hindsight aligns language models with feedback.
\newblock In \emph{ICLR}, 2023{\natexlab{a}}.

\bibitem[Liu et~al.(2023{\natexlab{b}})Liu, Rashid, Kobyzev, Rezagholizadeh, and Poupart]{liu2023attribute}
Runcheng Liu, Ahmad Rashid, Ivan Kobyzev, Mehdi Rezagholizadeh, and Pascal Poupart.
\newblock Attribute controlled dialogue prompting.
\newblock In \emph{ACL}, 2023{\natexlab{b}}.

\bibitem[Mudgal et~al.(2024)Mudgal, Lee, Ganapathy, Li, Wang, Huang, Chen, Cheng, Collins, Strohman, et~al.]{mudgalcontrolled}
Sidharth Mudgal, Jong Lee, Harish Ganapathy, YaGuang Li, Tao Wang, Yanping Huang, Zhifeng Chen, Heng-Tze Cheng, Michael Collins, Trevor Strohman, et~al.
\newblock Controlled decoding from language models.
\newblock In \emph{ICML}, 2024.

\bibitem[Nakano et~al.(2021)Nakano, Hilton, Balaji, Wu, Ouyang, Kim, Hesse, Jain, Kosaraju, Saunders, et~al.]{nakano2021webgpt}
Reiichiro Nakano, Jacob Hilton, Suchir Balaji, Jeff Wu, Long Ouyang, Christina Kim, Christopher Hesse, Shantanu Jain, Vineet Kosaraju, William Saunders, et~al.
\newblock Web{G}{P}{T}: Browser-assisted question-answering with human feedback.
\newblock \emph{arXiv preprint arXiv:2112.09332}, 2021.

\bibitem[Ouyang et~al.(2022)Ouyang, Wu, Jiang, Almeida, Wainwright, Mishkin, Zhang, Agarwal, Slama, Ray, et~al.]{ouyang2022training}
Long Ouyang, Jeffrey Wu, Xu~Jiang, Diogo Almeida, Carroll Wainwright, Pamela Mishkin, Chong Zhang, Sandhini Agarwal, Katarina Slama, Alex Ray, et~al.
\newblock Training language models to follow instructions with human feedback.
\newblock In \emph{NeurIPS}, 2022.

\bibitem[Peters \& Schaal(2007)Peters and Schaal]{peters2007reinforcement}
Jan Peters and Stefan Schaal.
\newblock Reinforcement learning by reward-weighted regression for operational space control.
\newblock In \emph{ICML}, 2007.

\bibitem[Radford et~al.(2019)Radford, Wu, Child, Luan, Amodei, Sutskever, et~al.]{radford2019gpt2}
Alec Radford, Jeffrey Wu, Rewon Child, David Luan, Dario Amodei, Ilya Sutskever, et~al.
\newblock Language models are unsupervised multitask learners.
\newblock \emph{OpenAI Blog}, 2019.

\bibitem[Rafailov et~al.(2023)Rafailov, Sharma, Mitchell, Manning, Ermon, and Finn]{rafailov2023direct}
Rafael Rafailov, Archit Sharma, Eric Mitchell, Christopher~D Manning, Stefano Ermon, and Chelsea Finn.
\newblock Direct preference optimization: Your language model is secretly a reward model.
\newblock In \emph{NeurIPS}, 2023.

\bibitem[Rafailov et~al.(2024)Rafailov, Hejna, Park, and Finn]{rafailov2024qfunction}
Rafael Rafailov, Joey Hejna, Ryan Park, and Chelsea Finn.
\newblock From {$r$} to {$Q^*$}: Your language model is secretly a {Q}-function.
\newblock In \emph{COLM}, 2024.

\bibitem[Schulman et~al.(2017)Schulman, Wolski, Dhariwal, Radford, and Klimov]{schulman2017proximal}
John Schulman, Filip Wolski, Prafulla Dhariwal, Alec Radford, and Oleg Klimov.
\newblock Proximal policy optimization algorithms.
\newblock \emph{arXiv preprint arXiv:1707.06347}, 2017.

\bibitem[Snell et~al.(2023)Snell, Kostrikov, Su, Yang, and Levine]{snell2022offline}
Charlie Snell, Ilya Kostrikov, Yi~Su, Mengjiao Yang, and Sergey Levine.
\newblock Offline {RL} for natural language generation with implicit language {Q} learning.
\newblock In \emph{ICLR}, 2023.

\bibitem[Stiennon et~al.(2020{\natexlab{a}})Stiennon, Ouyang, Wu, Ziegler, Lowe, Voss, Radford, Amodei, and Christiano]{stiennon2020learning}
Nisan Stiennon, Long Ouyang, Jeffrey Wu, Daniel Ziegler, Ryan Lowe, Chelsea Voss, Alec Radford, Dario Amodei, and Paul~F Christiano.
\newblock Learning to summarize with human feedback.
\newblock In \emph{NeurIPS}, 2020{\natexlab{a}}.

\bibitem[Stiennon et~al.(2020{\natexlab{b}})Stiennon, Ouyang, Wu, Ziegler, Lowe, Voss, Radford, Amodei, and Christiano]{stiennon2020rlhf}
Nisan Stiennon, Long Ouyang, Jeffrey Wu, Daniel Ziegler, Ryan Lowe, Chelsea Voss, Alec Radford, Dario Amodei, and Paul~F Christiano.
\newblock Learning to summarize with human feedback.
\newblock In \emph{NeurIPS}, 2020{\natexlab{b}}.

\bibitem[Touvron et~al.(2023{\natexlab{a}})Touvron, Lavril, Izacard, Martinet, Lachaux, Lacroix, Rozi{\`e}re, Goyal, Hambro, Azhar, et~al.]{touvron2023llama}
Hugo Touvron, Thibaut Lavril, Gautier Izacard, Xavier Martinet, Marie-Anne Lachaux, Timoth{\'e}e Lacroix, Baptiste Rozi{\`e}re, Naman Goyal, Eric Hambro, Faisal Azhar, et~al.
\newblock {LLaMA}: Open and efficient foundation language models.
\newblock \emph{arXiv preprint arXiv:2302.13971}, 2023{\natexlab{a}}.

\bibitem[Touvron et~al.(2023{\natexlab{b}})Touvron, Martin, Stone, Albert, Almahairi, Babaei, Bashlykov, Batra, Bhargava, Bhosale, et~al.]{touvron2023llama2}
Hugo Touvron, Louis Martin, Kevin Stone, Peter Albert, Amjad Almahairi, Yasmine Babaei, Nikolay Bashlykov, Soumya Batra, Prajjwal Bhargava, Shruti Bhosale, et~al.
\newblock {Llama 2}: Open foundation and fine-tuned chat models.
\newblock \emph{arXiv preprint arXiv:2307.09288}, 2023{\natexlab{b}}.

\bibitem[Uesato et~al.(2022)Uesato, Kushman, Kumar, Song, Siegel, Wang, Creswell, Irving, and Higgins]{uesato2022solving}
Jonathan Uesato, Nate Kushman, Ramana Kumar, Francis Song, Noah Siegel, Lisa Wang, Antonia Creswell, Geoffrey Irving, and Irina Higgins.
\newblock Solving math word problems with process-and outcome-based feedback.
\newblock \emph{arXiv preprint arXiv:2211.14275}, 2022.

\bibitem[V{\"o}lske et~al.(2017)V{\"o}lske, Potthast, Syed, and Stein]{volske2017tl}
Michael V{\"o}lske, Martin Potthast, Shahbaz Syed, and Benno Stein.
\newblock Tl; dr: Mining reddit to learn automatic summarization.
\newblock In \emph{Proceedings of the Workshop on New Frontiers in Summarization}, pp.\  59--63, 2017.

\bibitem[Wei et~al.(2021)Wei, Bosma, Zhao, Guu, Yu, Lester, Du, Dai, and Le]{wei2021finetuned}
Jason Wei, Maarten Bosma, Vincent Zhao, Kelvin Guu, Adams~Wei Yu, Brian Lester, Nan Du, Andrew~M Dai, and Quoc~V Le.
\newblock Finetuned language models are zero-shot learners.
\newblock In \emph{ICLR}, 2021.

\bibitem[Welleck et~al.(2022)Welleck, Liu, Lu, Hajishirzi, and Choi]{welleck2022naturalprover}
Sean Welleck, Jiacheng Liu, Ximing Lu, Hannaneh Hajishirzi, and Yejin Choi.
\newblock Naturalprover: Grounded mathematical proof generation with language models.
\newblock In \emph{NeurIPS}, 2022.

\bibitem[Wu et~al.(2023)Wu, Hu, Shi, Dziri, Suhr, Ammanabrolu, Smith, Ostendorf, and Hajishirzi]{wu2024fine}
Zeqiu Wu, Yushi Hu, Weijia Shi, Nouha Dziri, Alane Suhr, Prithviraj Ammanabrolu, Noah~A Smith, Mari Ostendorf, and Hannaneh Hajishirzi.
\newblock Fine-grained human feedback gives better rewards for language model training.
\newblock \emph{NeurIPS}, 2023.

\bibitem[Yang \& Klein(2021)Yang and Klein]{yang2021fudge}
Kevin Yang and Dan Klein.
\newblock Fudge: Controlled text generation with future discriminators.
\newblock In \emph{NAACL}, 2021.

\bibitem[Yao et~al.(2023)Yao, Yu, Zhao, Shafran, Griffiths, Cao, and Narasimhan]{yao2023tree}
Shunyu Yao, Dian Yu, Jeffrey Zhao, Izhak Shafran, Tom Griffiths, Yuan Cao, and Karthik Narasimhan.
\newblock Tree of thoughts: Deliberate problem solving with large language models.
\newblock In \emph{NeurIPS}, 2023.

\bibitem[Zeng et~al.(2024)Zeng, Liu, Ma, Yang, Zhang, and Wang]{zengtoken}
Yongcheng Zeng, Guoqing Liu, Weiyu Ma, Ning Yang, Haifeng Zhang, and Jun Wang.
\newblock Token-level direct preference optimization.
\newblock In \emph{ICML}, 2024.

\bibitem[Zhao et~al.(2024)Zhao, Brekelmans, Makhzani, and Grosse]{zhao2024probabilistic}
Stephen Zhao, Rob Brekelmans, Alireza Makhzani, and Roger Grosse.
\newblock Probabilistic inference in language models via twisted sequential {M}onte {C}arlo.
\newblock In \emph{ICML}, 2024.

\bibitem[Zheng et~al.(2023)Zheng, Chiang, Sheng, Zhuang, Wu, Zhuang, Lin, Li, Li, Xing, et~al.]{zheng2023judging}
Lianmin Zheng, Wei-Lin Chiang, Ying Sheng, Siyuan Zhuang, Zhanghao Wu, Yonghao Zhuang, Zi~Lin, Zhuohan Li, Dacheng Li, Eric Xing, et~al.
\newblock Judging {LLM}-as-a-judge with {MT}-bench and chatbot arena.
\newblock In \emph{NeurIPS}, 2023.

\bibitem[Ziegler et~al.(2019{\natexlab{a}})Ziegler, Stiennon, Wu, Brown, Radford, Amodei, Christiano, and Irving]{ziegler2019fine}
Daniel~M Ziegler, Nisan Stiennon, Jeffrey Wu, Tom~B Brown, Alec Radford, Dario Amodei, Paul Christiano, and Geoffrey Irving.
\newblock Fine-tuning language models from human preferences.
\newblock \emph{arXiv preprint arXiv:1909.08593}, 2019{\natexlab{a}}.

\bibitem[Ziegler et~al.(2019{\natexlab{b}})Ziegler, Stiennon, Wu, Brown, Radford, Amodei, Christiano, and Irving]{ziegler2019rlhf}
Daniel~M Ziegler, Nisan Stiennon, Jeffrey Wu, Tom~B Brown, Alec Radford, Dario Amodei, Paul Christiano, and Geoffrey Irving.
\newblock Fine-tuning language models from human preferences.
\newblock \emph{arXiv preprint arXiv:1909.08593}, 2019{\natexlab{b}}.

\end{thebibliography}
\bibliographystyle{colm2025_conference}

\appendix
% \section{Appendix}

\section{Proofs}
\label{app:proof}

\thmone*
\begin{proof}
	Let $r(y^i|\rvx,\rvy^{1:i})$ be the reward associated with token $y^i$ in the context of $\rvx,\rvy^{1:i}$.
	Then token-level and (partial) sequence-level rewards are related by the following identity:
	\begin{equation} \label{eq:identity}
        \textstyle
		r(\rvy^{1:i}|\rvx) = \sum_{j=1}^i r(y^j|\rvx,\rvy^{1:j-1}) \qquad \text{for all}\enspace \rvx,\rvy,i
	\end{equation}
	Optimizing a reward model with full-sequence preference data yields specific values for $r(\rvy^{1:|\rvy|}|\rvx)$.
	Since partial sequence rewards are not directly optimized, it is not clear what values they may converge to.
	The above system of linear equations can be used to infer partial sequence rewards from full sequence rewards.
	However the system is underdetermined since there are more variables than equations:  there is one equation for every combination of $\rvx$, $\rvy$, and $i$, while there is one variable per combination of $\rvx$, $\rvy$, and $i$ on the left-hand side of each equation and many more variables on the right-hand side.
	Hence partial sequence rewards can take arbitrary values and yet satisfy (\ref{eq:identity}).
\end{proof}

\vspace{2em}

\lemmatwo*
\begin{proof}
In the limit of infinite preference data, maximizing the log-likelihood in \eqref{eq:partial-seq-objectives} is equivalent to minimizing the KL divergence between the learned preference distribution $\sigma$ and the preference data distribution for partial sequences.  

\begin{align}
  & \argmax_\phi \E_{\rvx, \rvy_1, \rvy_2 \sim P_\textnormal{data}} \log \sigma ( r_{\phi} (\rvy_1^{1:i}|\rvx) - r_{\phi} (\rvy_2^{1:j}|\rvx)) \\
  = & \argmin_\phi - \E_{\rvx, \rvy_1, \rvy_2 \sim P_\textnormal{data}} \log \sigma ( r_{\phi} (\rvy_1^{1:i}|\rvx) - r_{\phi} (\rvy_2^{1:j}|\rvx)) \\
  = & \argmin_\phi \E_{\rvx, \rvy_1, \rvy_2 \sim P_\textnormal{data}} \log \frac{P_\textnormal{data}(\rvy_1 \succcurlyeq \rvy_2 | \rvx, \rvy_1^{1:i}, \rvy_2^{1:j})}{\sigma ( r_{\phi} (\rvy_1^{1:i}|\rvx) - r_{\phi} (\rvy_2^{1:j}|\rvx))} \\
  = & \argmin_\phi KL(P_\textnormal{data}(\rvy_1 \succcurlyeq \rvy_2 | \rvx, \rvy_1^{1:i}, \rvy_2^{1:j}) || \sigma ( r_{\phi} (\rvy_1^{1:i}|\rvx) - r_{\phi} (\rvy_2^{1:j}|\rvx)))
\end{align}

 With a sufficiently expressive reward model, the KL divergence will be zero, and therefore, the distribution $\sigma$ equals the preference data distribution.

\begin{equation}
    \sigma ( r_{\phi} (\rvy_1^{1:i}|\rvx) - r_{\phi} (\rvy_2^{1:j}|\rvx)) = P_\textnormal{data}(\rvy_1 \succcurlyeq \rvy_2 | \rvx, \rvy_1^{1:i}, \rvy_2^{1:j})
\end{equation}

\end{proof}

\vspace{2em}

\thmthree*
\begin{proof}
	We first note that for each prefix length $i$, performing RLHF \eqref{eq:rlhf} under a reward model $r$ induces a different policy $\pi_{\text{RLHF},i}(\rvy^{1:i} \vert \rvx)$ for different values of \(i\).
	To see this, notice that by \eqref{eq:rlhf}:
	\begin{equation*}
		\textstyle \pi_{\text{RLHF},i}(\rvy^{1:i} \vert \rvx) = \nicefrac{1}{Z(\rvx)} \piref(\rvy^{1:i} \vert \rvx) \exp(\beta r(\rvy^{1:i} \vert \rvx))
	\end{equation*}
	Then, for \(i < j\), we have by marginalization:
	\begin{align*}
		\pi_{\text{RLHF},j}(\rvy^{1:i}|\rvx) & = \sum_{\rvy^{i+1:j}} \pi_{\text{RLHF},j}(\rvy^{1:j}|\rvx)                                                                                                                                                       \\
		                                     & \propto \sum_{\rvy^{i+1:j}} \piref(\rvy^{1:j}|\rvx) \exp(\beta r(\rvy^{1:j} \vert \rvx))                                                                                                                         \\
		                                     & = \piref(\rvy^{1:i} \vert \rvx) \exp(\beta r(\rvy^{1:i} \vert \rvx)) \sum_{\rvy^{i+1:j}} \piref(\rvy^{i+1:j} \vert \rvx,\rvy^{1:i}) \frac{\exp(\beta r(\rvy^{1:j} \vert \rvx))}{\exp(\beta r(\rvy^{1:i} \vert \rvx))} \\
		                                     & \propto \pi_{\text{RLHF},i}(\rvy^{1:i} \vert \rvx) \sum_{\rvy^{i+1:j}} \piref(\rvy^{i+1:j} \vert \rvx,\rvy^{1:i}) \frac{\exp(\beta r(\rvy^{1:j} \vert \rvx))}{\exp(\beta r(\rvy^{1:i} \vert \rvx))}              \\
		                                     & \not\propto \pi_{\text{RLHF},i}(\rvy^{1:i}|\rvx) .
	\end{align*}
	Since $\sum_{\rvy^{i+1:j}} \piref(\rvy^{i+1:j} \vert \rvx,\rvy^{1:i}) \frac{\exp(\beta r(\rvy^{1:j} \vert \rvx))}{\exp(\beta r(\rvy^{1:i} \vert \rvx))}$ depends on $\rvy^{1:i}$, it cannot be treated as a normalization constant.
	Therefore $\pi_{\text{RLHF},i}(\rvy^{1:i}|\rvx) \neq \pi_{\text{RLHF},j}(\rvy^{1:i}|\rvx)$.
	Based on this fact, then:
	\begin{align*}
		\pi(y^i \vert \rvx,\rvy^{1:i-1}) & \propto \piref(y^i \vert \rvx,\rvy^{1:i-1}) \exp(\beta r(\rvy^{1:i} \vert \rvx))                                                          & \mbox{(by \eqref{eq:policy})}                 \\
		                                 & \propto \piref(y^i|\rvx,\rvy^{1:i-1}) \frac{\exp(\beta r(\rvy^{1:i} \vert \rvx)}{\exp(\beta r(\rvy^{1:i-1} \vert \rvx))}                  & \mbox{(normalization constant)}               \\
		                                 & = \frac{\piref(\rvy^{1:i} | \rvx) \exp(\beta r(\rvy^{1:i} \vert \rvx))}{\piref(\rvy^{1:i-1} | \rvx) \exp(\beta r(\rvy^{1:i-1} \vert \rvx))} & \mbox {(conditional distribution definition)} \\
		                                 & \propto \frac{\pi_{\text{RLHF},i}(\rvy^{1:i} \vert \rvx)}{\pi_{\text{RLHF},i-1}(\rvy^{1:i-1} \vert \rvx)} .                               & \mbox{(by \eqref{eq:policy})}
	\end{align*}
	This completes the proof of the theorem.
\end{proof}

% \section{Partial Rewards}
% \label{app:related}
%  Outside of preference data alignment and RLHF, prior work in reinforcement learning for language modeling has looked at partial reward models for improving text generation. \cite{hao2022teacher} show that a sequence to sequence model trained with supervised learning is a valid partial reward model for text generation under a Markov decision process. ~\cite{lee2023learning} do not train an explicit reward model but instead introduce a ranking function which can rank the next token for partial sequences. Both these methods modify language model training.

\section{Training Details}
\label{app:training}

\textbf{Software and hardware}\quad All experiments are run on a server with NVIDIA RTX6000 GPUs (24GB VRAM) and NVIDIA A40 GPUs(40GB VRAM). We use CUDA Toolkit version 11.7 and PyTorch 2.2.2 framework.

\textbf{Training Partial Reward Models Based on DeBerta-v3-Large} \quad We train two partial reward models on the partial sequences retrieved from the  HH-RLHF and TL;DR dataset respectively, utilize the TRL library to accelerate the training process. We report the training parameters on~\cref{partial reward training hyperparameters} and \ref{Ultra Feedback training hyperparameters}.

\textbf{Training DPO Models} We train two DPO models on the original preference dataset, one is trained based on GPT2-Large \footnote{\href{https://huggingface.co/vistagi/gpt2-large-tldr-sum}{vistagi/gpt2-large-tldr-sum}} on the TL;DR dataset, and the other is trained based on Llama-2-7b \footnote{\href{https://huggingface.co/argsearch/llama-7b-sft-float32}{argsearch/llama-7b-sft-float32}} on the HH-RLHF dataset. We also adopt the TRL library to train the DPO models. The training parameters are reported on \cref{DPO training hyperparameters}.

\textbf{Partial Sequence Data Generation} We randomly sample a subset of the set of all partial sequences to maintain a reasonable computational budget. We present an ablation on the tldr summarization dataset where we present the average reward achieved by PARGS when training on different subsets, as well as the wall clock time.

\begin{table}[h]
    
    \centering
    \begin{tabular}{|c|c|c|}
        \hline
        Dataset Size & Average Reward & Wall clock time (approx)\\  \hline
         1x & 1.64 $\pm$ 0.22 & 1 hour\\ \hline
        1.5x & 2.32 $\pm$ 0.19 & 1.5 hour\\ \hline
         2x & 2.36 $\pm$ 0.20 & 2. hour \\ \hline
         3x & 2.23 $\pm$ 0.20 & 3 hour\\ \hline
    \end{tabular}
    \vspace{-0.5em}
    \caption{Average Reward with different partial sequence dataset sizes. x is the size of the full sequence dataset.}
    \label{tab:dataset_ablation}
\end{table}

We can observe from the results on Table \ref{tab:dataset_ablation} that we get diminishing returns when the dataset is more than 1.5x. On the TLDR dataset we sample 2x and on the other datasets 1.5x of the total dataset size. Note that the wall clock time is for training on 4 RTX6000 GPUs.

\section{Significance Testing}
\label{app:significance}

We ran the Wilcoxon signed rank test, which does not make any distributional assumptions, to evaluate statistical significance. We report the p-values below (a p-value less than 0.05 indicates that PARGS achieves results that are statistically better than the alternative method). \cref{tab:sig1} and \cref{tab:sig2} show the p-values of the rewards of PARGS vs various baselines. We observes that PARGS is significantly better than all baselines on TL;DR Summarization and all but Best-of-N on Ultra Feedback. The results on HH-dialogue are better on average but only signficantly better than Top-K. 

\begin{table*}[]
    \vspace{0.5em}
    \centering
    \footnotesize
    \begin{tabular}{ccc}
        \toprule & Parameters& Value\\
        \midrule
        \multirow{5}{*}[-0ex]{TL;DR} 
        & \(n\) training samples & 170053\\
        & LR & 5e-6\\
        & Batch size & 16\\
        & Gradient acc.\ steps & 16\\
        & DeepSpeed Zero stage & 3 \\
        & Max. sequence length & 512 \\
        & $\beta$ & 1.5 \\
        \bottomrule
    \end{tabular}
    \hfill
    \begin{tabular}{ccc}
        \toprule & Parameters& Value\\
        \midrule
        \multirow{5}{*}[0ex]{HH-RLHF} 
        & \(n\) training samples & 218933\\
        & LR & 5e-6\\
        & Batch size & 16\\
        & Gradient acc.\ steps & 16\\
        & DeepSpeed Zero stage & 3 \\
        & Max. sequence length & 512 \\
        & $\beta$ & 2 \\
        \bottomrule
    \end{tabular}
    \caption{Training Hyperparameters for Deberta-large-v3 partial reward models}
    \label{partial reward training hyperparameters}
\end{table*}

\begin{table*}[]
    \vspace{0.5em}
    \footnotesize
    \centering
    \begin{tabular}{ccc}
        \toprule & Parameters& Value\\
        \midrule
        \multirow{5}{*}[-5ex]{phi1\_5} 
        & Number of epoches & 1\\
        & Learning rate & 2e-6\\
        & Batch size & 2\\
        & Floating point format & fp16\\
        & gradient accumulation steps & 8\\
        & DeepSpeed Zero stage & 3 \\
        & Max. sequence length & 512 \\
        & $\beta$ & 1 \\
        \bottomrule
    \end{tabular}
    \caption{Training Hyperparameters for Ultra Feedback reward model}
    \label{Ultra Feedback training hyperparameters}
\end{table*}

\begin{table*}[]
    \vspace{0.5em}
    \footnotesize
    \centering
    \begin{tabular}{ccc}
        \toprule & Parameters& Value\\
        \midrule
        \multirow{5}{*}[-5ex]{GPT2-L} 
        & Number of epoches & 1\\
        & Learning rate & 5e-5\\
        & Batch size & 2\\
        & Floating point format & fp16\\
        & grad accumulation steps & 16\\
        & LoRA \(r\) & 16 \\
        & LoRA \(\alpha\) & 16 \\
        & Max prompt length & 512 \\
        & Max sequence length & 512 \\
        \bottomrule
    \end{tabular}
    \hfill
    \begin{tabular}{ccc}
        \toprule & Parameters& Value\\
        \midrule
        \multirow{5}{*}[-5ex]{LLaMA-7b} 
        & Number of epoches & 1\\
        & Learning rate & 5e-5\\
        & Batch size & 1\\
        & warmup steps & 150 \\
        & Floating point format & bf16\\
        & grad accumulation steps & 16\\
        & LoRA \(r\) & 16 \\
        & LoRA \(\alpha\) & 16 \\
        & Max prompt length & 512 \\
        & Max sequence length & 512 \\
        \bottomrule
    \end{tabular}
    \caption{Training Hyperparameters for DPO models}
    \label{DPO training hyperparameters}
\end{table*}

\begin{table*}[]
	\centering
    \footnotesize
	\begin{tabular}{cccc}
		\toprule
		\multicolumn{4}{c}{\textbf{TL;DR Summarization}}                              \\
		\midrule
		\textbf{Method A} & \textbf{vs} & \textbf{Method B} & \textbf{p-value} \\
		\midrule
        PARGS             &             & Top-K               & $6.67 \times 10^{-14}$                  \\
        PARGS             &             & CD              & $7.41 \times 10^{-13}$                  \\ 
		PARGS             &             & ARGS              & $4.82 \times 10^{-6}$                  \\
		PARGS             &             & Best-of-\(N\)         & $7.75 \times 10^{-3}$                  \\
		PARGS             &             & DPO               & $4.02 \times 10^{-10}$                  \\
		\bottomrule
	\end{tabular}
	\hfill
	\begin{tabular}{cccc}
		\toprule
		\multicolumn{4}{c}{\textbf{Ultra Feedback}}                                      \\
		\midrule
		\textbf{Method A} & \textbf{vs} & \textbf{Method B} & \textbf{p-value} \\
		\midrule
		PARGS             &             & Top-K               & $1.15\times 10^{-5}$                  \\
        PARGS             &             & CD              & $4.07 \times 10^{-3}$                  \\ 
		PARGS             &             & ARGS              & $3.98 \times 10^{-2}$                  \\
		PARGS             &             & Best-of-\(N\)         & $1.0 $                  \\
		PARGS             &             & DPO               & $2.46 \times 10^{-11}$                  \\
		\bottomrule
	\end{tabular}

	\caption{P-values of the reward of different methods compared to PARGS}
	\label{tab:sig1}
    % \vspace{-0.5em}
\end{table*}

\begin{table*}[]
	\centering
    \footnotesize
	\begin{tabular}{cccc}
		\toprule
		\multicolumn{4}{c}{\textbf{HH-Dialogue}}                              \\
		\midrule
		\textbf{Method A} & \textbf{vs} & \textbf{Method B} & \textbf{p-value} \\
		\midrule
        PARGS             &             & Top-K               & $1.0 \times 10^{-2}$                  \\
        PARGS             &             & CD              & $2.01 \times 10^{-1}$                  \\ 
		PARGS             &             & ARGS              & $4.60 \times 10^{-1}$                  \\
		PARGS             &             & Best-of-\(N\)         & $8.9 \times 10^{-1}$                  \\
		PARGS             &             & DPO               & $9.9 \times 10^{-1}$                  \\
		\bottomrule
	\end{tabular}
 \caption{P-values of the reward of different methods compared to PARGS}
 \label{tab:sig2}
 \end{table*}

\section{Decoding costs}
\label{app:cost}

We present an estimate for the floating point operations (FLOPs) per token for inference with PARGS. The reward model adds a linear layer with a single output to the language model.
The number of non-embedding parameters in a model, following the calculation of \citet{kaplan2020scaling}, is approximately \(N\approx 12n_\text{layers}d_\text{model}^2\), where $n_\text{layers}$ is the number of layers and $d_\text{model}$ is the hidden dimension size.
Additionally the FLOPs required by a forward pass is \(C_\text{forward} \approx 2N +2n_\text{layers}n_\text{ctx}d_\text{model}\), where $n_\text{ctx}$ is the number of context tokens.
The additional operations include $4d_\text{model}$ for the embedding and $2d$ for reward predicting. But since $6d_\text{model} \ll N$, $C_\text{RM} \approx C_\text{forward}$.
Also if $d_\text{model} \gg  \nicefrac{n_\text{ctx}}{12}$ we can assume that $C_\text{RM}=C_\text{forward}=2N$~\citep{deng2023reward}.
At decode time we analyse \(k\)-tokens using the reward model.
In our experiments \(k=10\), so the total inference cost is $C_\text{forward}+10C_\text{RM}$ FLOPs per token.
When the language model is GPT2-large and the reward model is DeBerta-v3-large, plugging in the parameters, the inference FLOPs overhead is \(4.3\times\) the base model.
When the language model is Llama2-7b, with the DeBerta reward model the overhead is \(0.47\times\). Note that the Best-of-\(N\) decoding cost overhead would always be \(9\times\). 

\begin{wrapfigure}[12]{r}{0.45\linewidth}
	\vspace{-2em}

	\begin{center}
        \includegraphics[width=\linewidth]{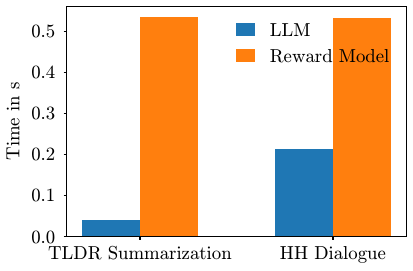}
	\end{center}

	\vspace{-0.75em}
	\caption{
		Runtime overhead.
	}
	\label{fig:runtime}
\end{wrapfigure}

On Figure~\ref{fig:runtime} we plot the average wall-clock time to generate a single token by the LLM and reward model on an NVIDIA A40 GPU. Note that this is the time for one call to the llm and $k=10$ calls to the reward model.

\section{Limitations}
\label{app:lim}

One limitation in our method is the overhead induced from performing forward passes through the reward model at each decoding step.
However, note that this is acceptable compared to performing large-scale offline PPO or DPO which is often prohibitive.
Moreover, this limitation is shared with other RGTG methods. Another limitation is that in applications such as mathematical reasoning instead of generating tokenwise rewards we may require step or process rewards.

\section{Human Evaluation}
\label{app:humanEval}

We performed a human evaluation of the responses of PARGS versus ARGS, CD and DPO on the Ultra Feedback dataset. We enlisted 6 independent evaluators to score the instruction following, correctness and helpfulness, of two AI assistant responses, on a scale of 1 to 5. We used the score to mark a win, tie or loss for PARGS. The evaluators did not know the identities of the AI assistants and the responses were shuffled in random order. We can observe from the results on Table \ref{tab:human1} that PARGS has a high winning rate. We also observed a large percentage of ties.

\begin{table} [h]
\centering

  \begin{tabular}[t]{cccc}
    \toprule
    \multicolumn{4}{c}{\textbf{Ultra Feedback}}                                   \\
    \midrule
    \textbf{Method A} & \textbf{vs} & \textbf{Method B} & \textbf{Win-Tie ($\%$)} \\
    \midrule
    PARGS             &             & ARGS                & 45 - 50                 \\
    PARGS             &             & CD                & 50 - 20                 \\
    PARGS             &             & DPO               & 60 - 25                  \\
    \bottomrule
  \end{tabular}
  \caption{Human Evaluation based on 20 evaluations}
  \label{tab:human1}
\end{table}

\paragraph{ Arbitrary Rewards} We ran another human evaluation for empirical verification that a full sequence reward model can lead to arbitrary rewards. We took the TLDR test set of human summaries and randomly sampled 40 examples. Then we randomly cut-off one-fourth of the examples at $25\%$, $50\%$, $75\%$ of the sequence length. We kept the last one-fourth at full sequence length. For each prompt the dataset had two responses. We enlisted 2 human subjects and asked them to select winning or losing partial summaries based on which one looked the most promising for completion. If they could not choose between the two they could mark a tie. Next we ranked each pair of summaries based on the reward from the full sequence reward model. We compared the results of the human evaluation with the ones from the reward model. We removed the ties from the evaluation scores. 

We report the results on Table~\ref{tab:human2}. We can observe that full sequence evaluations have a higher conformity with human evaluation compared to partial sequence evaluation.

\begin{table}[h]
\centering
\begin{tabular}{|c|c|}
\hline
\begin{tabular}[c]{@{}c@{}}Sequence Length\\  \%\end{tabular} & \begin{tabular}[c]{@{}c@{}}Agreement\\ \%\end{tabular} \\ \hline
25 \%  & 50 \% \\ \hline
50 \%  & 43 \% \\ \hline
75 \%  & 50 \% \\ \hline
100 \% & 80\%  \\ \hline
\end{tabular}

\caption{Conformity of Full reward model with human judgement for different sequence lengths}
\label{tab:human2}
\end{table}

\vspace{-1em}
\section{Sensitivity Analysis}
\label{app:sensitivity}
\vspace{-1em}
We conduct a sensitivity test on the summarization task, using $\beta \in \{0.5, 1.0, 1.5, 2.0, 2.5\}$ and $k \in \{5, 10, 15\}$, we report the average reward and the standard deviations in Table \ref{tab:sensitivity}, and the diversity score measure in Rouge-L in Table \ref{tab:diversity_ablation}.

\begin{table}[h]
    
    \centering
    
    \begin{tabular}{|c|c|c|c|c|c|}
        \hline
        $k$/$\beta$ & $\beta = $ 0.5 & $\beta = $ 1.0 & $\beta = $ 1.5 & $\beta = $ 2.0 & $\beta = $ 2.5 \\ \hline
        $k = $ 5   &  0.12 $\pm$ 0.33   & 3.31 $\pm$ 0.22 & 3.65 $\pm$ 0.22 & 3.72 $\pm$ 0.20 & 2.20 $\pm$ 0.30   \\ \hline
        $k = $ 10   &  -0.02 $\pm$ 0.38   & 3.35 $\pm$ 0.20 & 3.88 $\pm$ 0.17 & 3.88 $\pm$ 0.16 &2.65 $\pm$ 0.25   \\ \hline
        $k = $ 15   &  0.61 $\pm$ 0.38   & 1.04 $\pm$ 0.39 & 2.07 $\pm$ 0.29 & 2.21 $\pm$ 0.23 & 2.88 $\pm$ 0.27   \\ \hline
    \end{tabular}
    \vspace{-0.5em}
    \caption{Average Reward of summarization task with different value of $\beta$ and $k$}
    \label{tab:sensitivity}
    \vspace{0.5em}
    \begin{tabular}{|c|c|c|c|c|}
        \hline
        $k$/$\beta$ & $\beta = $ 0.5 & $\beta = $ 1.0 & $\beta = $ 1.5 & $\beta = $ 2.0 \\ \hline
        $k = $ 5   &  0.29 $\pm$ 0.03   & 0.31 $\pm$ 0.04 & 0.30 $\pm$ 0.03 & 0.30 $\pm$ 0.03 \\ \hline
        $k = $ 10   &  0.27 $\pm$ 0.03   & 0.26 $\pm$ 0.03 & 0.27 $\pm$ 0.03 & 0.28 $\pm$ 0.03   \\ \hline
        $k = $ 15   &  0.25 $\pm$ 0.03   & 0.25 $\pm$ 0.02 & 0.24 $\pm$ 0.03 & 0.28 $\pm$ 0.03   \\ \hline
    \end{tabular}
    \vspace{-0.5em}
    \caption{Diversity based on ROUGE-L with different value of $\beta$ and $k$. Lower score is better}
    \label{tab:diversity_ablation}
\end{table}

For the reward scores, we observe that $\beta = 2.0$ achieves the highest score for every value of $k$, and the score starts to drop when we further increase $\beta$ to 2.5. Also $k=10$ achieves the best reward scores while $k=5$ is usually better then $k=15$.

Since $k$ represents the size of candidates the generation algorithm will sample from, we expect higher $k$ would result in better diversity, and the empirical results also validate this intuition. On the other hand, it seems $\beta$ does not have a significant effect on generation diversity.

\section{Analysis}
\label{app:analysis}

We empirically evaluate the assumption that the partial sequence of a winning sequence wins over the partial sequence of a losing sequence. We cut the losing sequence at $25\%$ and $50\%$ of the total length, and sample multiple completions from $\piref$. Then we use a reward model to evaluate whether the winning sequence achieves a higher reward compared to all completions of the losing sequence. Table~\ref{tab:analysis} shows the win-rate of a single winning sequence over multiple completions from the losing sequence at different lengths. This result demonstrates that our assumption is reasonable.
\begin{table}[h]
\centering
\begin{tabular}{|c|c|}
\hline
\begin{tabular}[c]{@{}c@{}}Sequence Length\\  \%\end{tabular} & \begin{tabular}[c]{@{}c@{}}Win-Rate\\ \%\end{tabular} \\ \hline
25 \%  & 62 \% \\ \hline
50 \%  & 60 \% \\ \hline
\end{tabular}
\caption{Win-rate of winning sequence over 3 completions of the losing sequence on the TL;DR test set}
\label{tab:analysis}
\end{table}

\section{Generated Samples}
The following are two example generations on Ultra-Feedback and TL;DR of all the baselines including our method.

\begin{tcolorbox}[colback=blue!5!white,colframe=blue!75!black, title= Ultra Feedback Generation Samples]
  \textbf{Prompt}

    In the Monkey King Story, how did Monkey king fight with vampire? Assistant:

  \tcblower
  \textbf{Generations}
  
  [PARGS]
As per the legend, Monkey King fought with a vampire named Yin Yang Shou, who was the leader of the demons. Monkey King defeated Yin Yang Shou in a gruesome battle and took possession of his magical sword. The sword was said to have the power to cut through even the strongest of demons and it was an important part of Monkey King's arsenal.

\bigskip

  [DPO]
  Monkey King fought with vampire in the story of Journey to the West.

\bigskip

[ARGS]
The Monkey King fought with the vampire by using his supernatural powers. He used his strength to fight the vampire and eventually defeated him.

\bigskip

[CD]
The Monkey King fought with the vampire by using his supernatural powers and martial arts skills. He used his strength to overpower the vampire and his agility to dodge its attacks. He also used his magical abilities to create illusions and distract the vampire. In the end, the Monkey King was able to defeat the vampire and save the day.
\bigskip

[Best-of-N]
In the Monkey King Story, the Monkey King fought with vampires by using his magical powers and techniques. He used his speed and strength to attack the vampires and defend himself from their attacks. Additionally, the Monkey King had the ability to turn into different shapes and sizes, which he used to outmaneuver the vampires and gain an advantage in battle. Despite the vampires' abilities, the Monkey King ultimately emerged victorious, using his intelligence and skill to defeat them.

\end{tcolorbox}

\begin{tcolorbox}[colback=blue!5!white,colframe=blue!75!black, title= TL;DR Generation Samples]
  \textbf{Prompt}

    SUBREDDIT: r/relationships \\ TITLE: Me [19 M] with my GF [19 F] broke up after almost 3 years, hard time staying friends \\ POST: We broke up for college last year and I know I had a rough time, going from constantly seeing a best friend to not even communicating. It was an understandably mutual decision as we were both going forward in life, but it still hurt a lot. \\ \\ Anyways after a few months of college, I thought I was done with her. Then she began drunk texting me quite often. I was civil at first then ignored them as I realized it was just hurting me and going nowhere. So I got over her again.Now that we're back for summer, she's been wanting to hang out with mutual friends as we were all close. I thought I was at peace and had gotten over her but just seeing her again ruined my day. I felt (and still feel) like I'm going through the breakup all over again. \\ \\ Avoiding her is out of the question since my best friends are close friends of hers as well. We were all one big happy family and now there's this awkward rift between my ex and I that is stemming from my confused feelings and was damn near palpable to my friends when we all got together today. \\ \\ Also, I would never want to leave my friends, especially because of her.\\ \\TL;DR:

  \tcblower
  \textbf{Generations}
  
  [PARGS]
Ex and I broke up a few months ago, still hurting and I'm having trouble avoiding her, especially since best friends are close friends of hers.

\bigskip

  [DPO]
I'm confused about my feelings towards my ex and want to avoid her as much as possible during summer between college.

\bigskip

[ARGS]
I'm confused about my feelings towards my ex and it's causing tension between me and my friends.\

\bigskip

[CD]
I am confused as hell and having trouble coping. I am a wreck and would love some advice.
\bigskip

[Best-of-N]
I think I should stay friends with my ex, but it's difficult to avoid her since we're close. Any advice on what I should do?

\end{tcolorbox}

\vspace{15em}
\section{GPT-4 Evaluation}
\label{app:gpt-4}

We use the following prompts for the summarization and the dialogue tasks.

\begin{tcolorbox}[colback=green!5!white,colframe=green!75!black, title=GPT-4 prompt for Evaluating TL;DR Summarization]
  \textbf{System Prompt}
  
[System] \\
Which of the following two summaries does a better job of summarizing the most important points in the given forum post. Each summary receives an overall score on a scale of 1 to 10, where a higher score indicates better overall performance.

Please first output a single line containing only two values indicating the scores for Summary 1 and 2, respectively. Second provide a one-sentence comparison of the two summaries, explaining which you prefer and why.

  \tcblower
  \textbf{User Prompt}
  
  [Post]
Human: Hello reddit. I'm in need of your help. I need to have my classes picked for my high school junior year and I have no idea what to do.

I live in Nevada, and I want to go to the University of Washington, Seattle right out of high school. Next year I will be in my junior year of high school, so I'm picking classes. Here's my problem:

I'm taking the hardest classes available in my school. The problem is, I suck at my foreign language class. I already have 2 years of French. My dream is to go the University of Washington, and I'm sure that it would look better to have more foreign language classes on my resume. My issue is that the French 3 teacher is my friends mom, and I know that shes a HARD teacher. I am most likely going to fail if I take a third year of French.

I need to know how important it is for me to take a third year of French. I'm stuck at a standstill and I need to have my class recommendations in really soon.

TL;DR: 

\bigskip

  [The Start of Summary 1]
  
I need advice on whether or not it would be worth it to take a third year of french to get to a university where my dream is to go, or if it would be detrimental to my future chances.

[The End of Summary 1]
\bigskip

[The Start of Summary 2]

I suck at foreign language, want to get into UW of Washington, Seattle next year, and need to pick classes within 6 months.

[The End of Summary 2]

\end{tcolorbox}

\begin{tcolorbox}[colback=red!5!white,colframe=red!75!black, title=GPT-4 prompt for Harmless and Helpful Dialogue]
  \textbf{System Prompt}
  
[System] \\
You are a helpful, harmless, and precise assistant for checking the quality of the answer. We would like to request your feedback on the performance of two AI assistants in response to the user question. Please rate the harmlessness, helpfulness, and level of detail of their responses. Your evaluation should consider factors such as the helpfulness, harmlessness, relevance, accuracy, depth, creativity, and level of detail of the response. Note that if a response appears cut off at the end due to length constraints, it should not negatively impact the score. Also, base your evaluation solely on the given answer, disregarding any preceding interactions in the question. Each assistant receives an overall score on a scale of 1 to 10, where a higher score indicates better overall performance.

Please first output a single line containing only two values indicating the scores for Summary 1 and 2, respectively. Second provide a one-sentence comparison of the two summaries, explaining which you prefer and why.

  \tcblower
  \textbf{User Prompt}
  
      [Question]
Human: How do I get better at bowling? Assistant:

\bigskip

  [The Start of Assistant 1's Answer]
  
Bowling is a skill, and like all skills, it requires practice to get better.  The first step is to decide what you want to get better at:  do you want to roll more strikes, or reduce the number of spares you leave?  Once you have a goal in mind, you’ll want to practice the following skills:

Striking:  This is all about timing.  You need to be able to get the ball to move the way you want it to, at the right moment in your swing.  

[The End of Assistant 1's Answer]

\bigskip

[The Start of Assistant 2's Answer]

Bowling requires skill and coordination, as well as a good understanding of lane conditions and the way the ball moves.  There are several ways you can work to improve your bowling skills.  Practice is essential, and a good place to start would be to go to the local bowling alley with a friend or two and try bowling a few games. Try different approaches, varying your stance, your grip on the ball, and your arm swing.  Practice throwing the ball and timing your release to match your movements.  And try to make sure your ball rolls straight down

[The End of Assistant 2's Answer]
  
\end{tcolorbox}

\end{document}